\documentclass{article}

\usepackage[preprint,nonatbib]{neurips_2024}

\usepackage[utf8]{inputenc} 
\usepackage[T1]{fontenc}    
\usepackage[backref=page]{hyperref}       
\usepackage{url}            
\usepackage{booktabs}       
\usepackage{amsfonts}       
\usepackage{nicefrac}       
\usepackage{microtype}      
\usepackage{xcolor}         
\usepackage{graphicx}
\usepackage{xcolor}
\usepackage[ruled,vlined]{algorithm2e}
\usepackage[square,numbers]{natbib}
\usepackage{soul}
\usepackage{tikz}
\usepackage{amssymb}
\usepackage{nicefrac} 
\usepackage{amsmath}
\usepackage{breqn}
\usepackage[english]{babel}

\usepackage{wrapfig}
\usepackage{caption}
\usepackage{mathtools}
\usepackage{multirow}
\usepackage{algpseudocode}
\usepackage{comment}
\usepackage{subcaption}
\usepackage{listings}
\usepackage{float}
\usepackage{enumitem}
\usepackage{array}
\usepackage{threeparttable}

\newtheorem{theorem}{Theorem}
\newtheorem{property}{Property}
\newtheorem{proof}{Proof}[theorem]
\newtheorem{definition}{Definition} 

\newcolumntype{C}[1]{>{\centering\arraybackslash}m{#1}}

\usepackage{lipsum}
\title{VidModEx: Interpretable and Efficient Black Box Model Extraction for High-Dimensional Spaces}

\author{%
  Somnath Sendhil Kumar\textsuperscript{\dag} \And
  Yuvaraj Govindarajulu\textsuperscript{\ddag} \And
  Pavan Kulkarni\textsuperscript{\ddag} \And
  Manojkumar Parmar\textsuperscript{\ddag}
  \\ \\
  \textsuperscript{\dag} Microsoft Research, India.\\
  \textsuperscript{\ddag} AIShield, Bosch Global Software Technologies, Bangalore, India.\\ \texttt{\{yuvaraj.govindarajulu, pavan.kulkarni, manojkumar.parmar\}@bosch.com}
}

\begin{document}

\maketitle

\begin{abstract}
  In the domain of black-box model extraction, conventional methods reliant on soft labels or surrogate datasets struggle with scaling to high-dimensional input spaces and managing the complexity of an extensive array of interrelated classes. In this work, we present a novel approach that utilizes SHAP (SHapley Additive exPlanations) to enhance synthetic data generation. SHAP quantifies the individual contributions of each input feature towards the victim model's output, facilitating the optimization of an energy-based GAN towards a desirable output. This method significantly boosts performance, achieving a 16.45\% increase in the accuracy of image classification models and extending to video classification models with an average improvement of 26.11\% and a maximum of 33.36\% on challenging datasets such as UCF11, UCF101, Kinetics 400, Kinetics 600, and Something-Something V2. We further demonstrate the effectiveness and practical utility of our method under various scenarios, including the availability of top-k prediction probabilities, top-k prediction labels, and top-1 labels.
\end{abstract}

\section{Introduction}\label{sec:intro}

\begin{wrapfigure}{r}{.35\linewidth}
\begin{minipage}{\linewidth}
    \vspace{-6pt}
    \includegraphics[width=\linewidth, clip, trim=0cm 0cm 0cm 0cm]{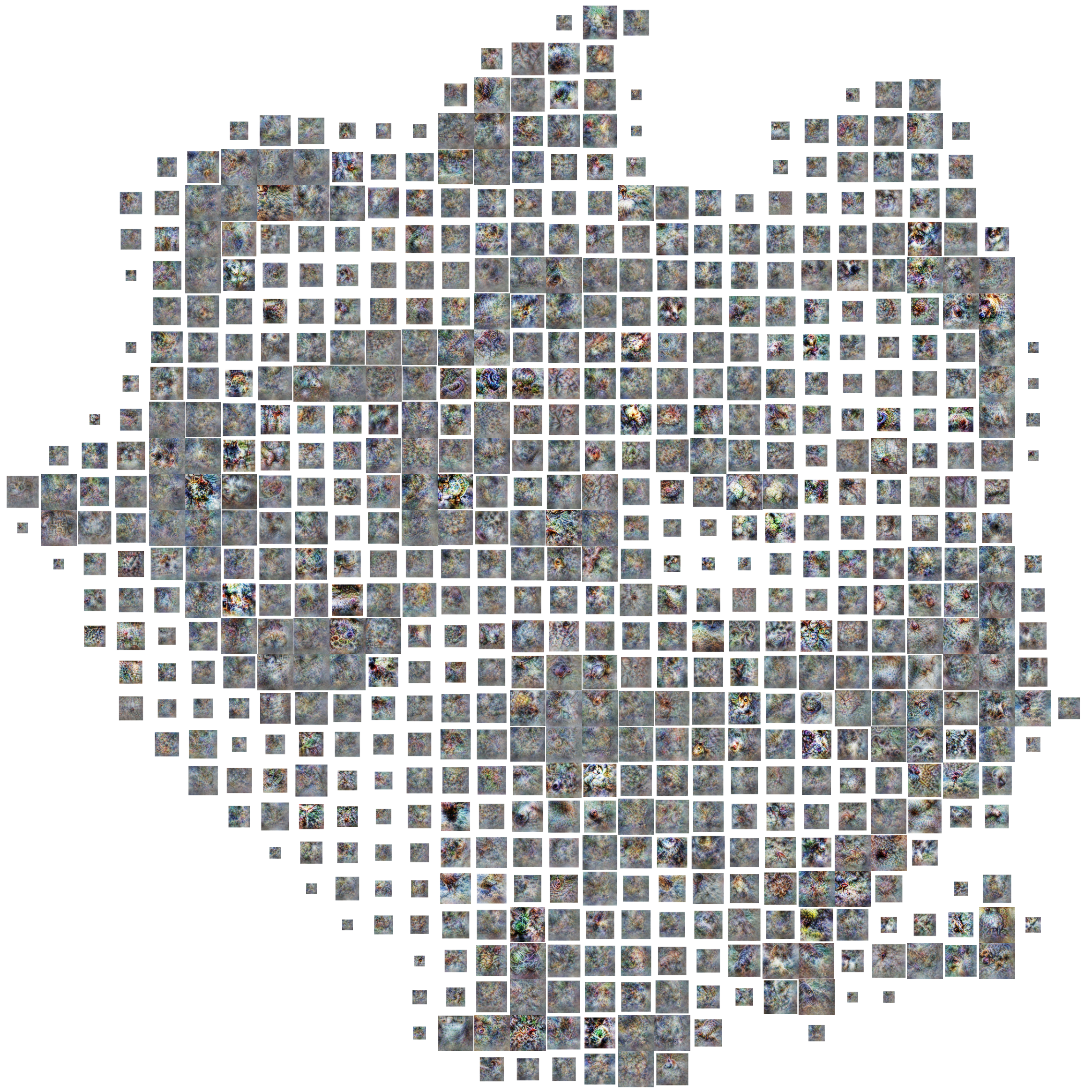}
     \captionof{figure}{\centering Act. Atlas for DFME\cite{truong2021data}}
    \label{fig:nonshaplucent}
    \vspace{-4pt}
\end{minipage}
\end{wrapfigure}

With the rise in MLaaS (Machine Learning as a Service), which performs tasks from minute levels \cite{azureaivision}, \cite{awscomputervision}\&\cite{edenai} to multitasking across domains \cite{gptvsystemcard}, \cite{covariantrfm1}; There has been a significant increase in model performance, correlating with their size and the ability to accommodate large input spaces. Previous model extraction attacks \cite{truong2021data},\cite{miura2021megex}, \cite{sanyal2022towards} \& \cite{wang2022blackdissector} have predominantly targeted small datasets such as MNIST and CIFAR, and at the best case scenario have achieved acceptable extraction accuracy on CIFAR-100, which are easily outperformed by current datasets and more robust models. Although there are studies scaling to large real-world models like \cite{carlini2024stealing}, these are specifically crafted for a target architecture or task, making a generalized approach challenging. 



On the contrary, some methods employ surrogate datasets \cite{truong2021data}, \cite{sanyal2022towards}, \cite{han2024exploring}, \cite{zhao2023extracting}, \cite{yan2022holistic} to train a substitute models, providing a prior about the target dataset. However, studies finding a balance \cite{han2024exploring} between surrogate and target datasets are limited in terms of scalability. With the affordable cost of hardware and increased services offering model fine-tuning for user data\cite{huggingfaceautotrain}, relying on surrogate datasets presents challenges in selecting the appropriate dataset. While every task in model extraction comes with its nuances, ranging from classification problems that might use soft labels or hard labels to top-k predictions/labels or top-1 prediction/label \cite{kazhdan2020marleme}, \cite{beetham2023dualstudent}, \cite{he2021model}, a generalized base approach can promote the development of more efficient and large-scale attacks. In this work, we limit ourselves to Vision Classifiers, but we do not exploit any specific architectural constraints or any discrepancy present in these tasks, thus maintaining an approach that is easily adaptable to other domains.

\begin{wrapfigure}{r}{.35\linewidth}
\begin{minipage}{\linewidth}
    \vspace{-8pt}
    \includegraphics[width=\linewidth, clip, trim=0cm 0cm 0cm 0cm]{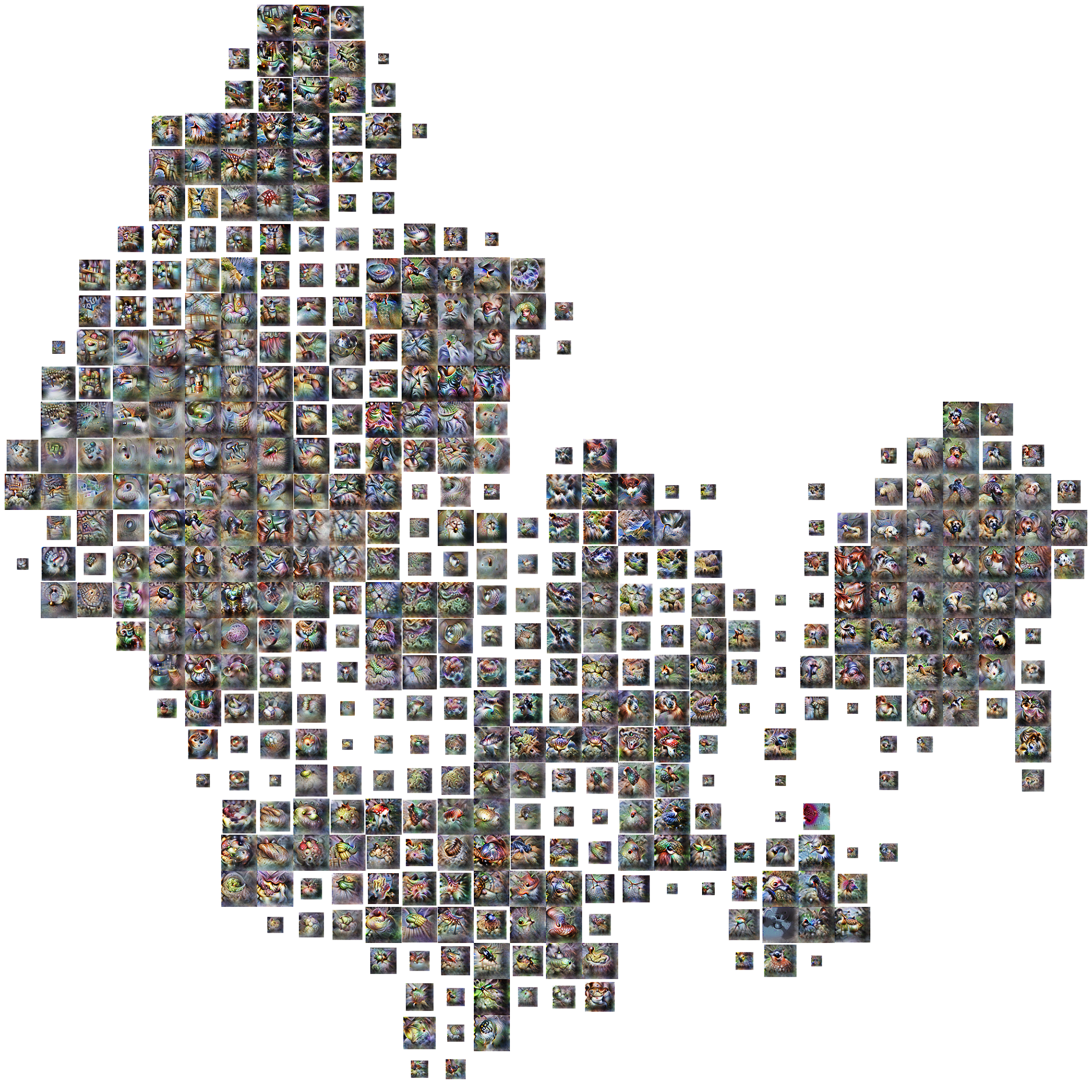}
    \captionof{figure}{\centering Act. Atlas for SHAP}
    \label{fig:shaplucent}
    
\end{minipage}
\end{wrapfigure}

\begin{figure}
\includegraphics[trim=0.3cm 0cm 0.3cm 0cm,clip,width=1.0\linewidth]{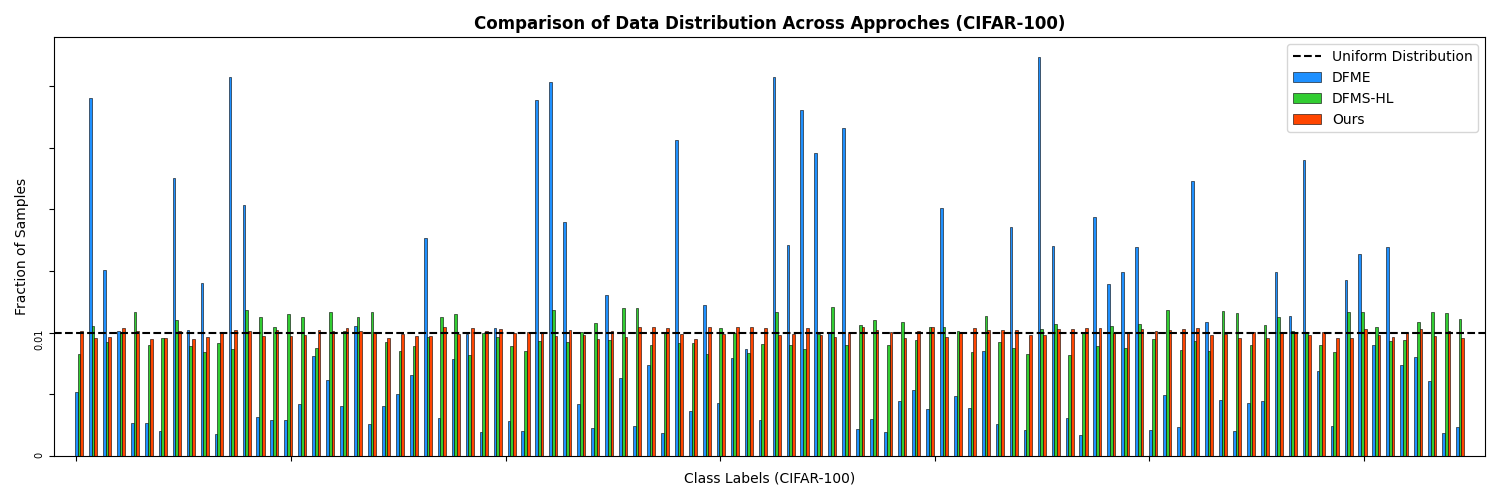}
\caption{Distribution based on Victim model prediction on generated samples for CIFAR 100}\label{fig:class_dist}
\vspace{-14pt}
\end{figure}

We employ SHAP\cite{lundberg2017unified}, an InterpretableAI Algorithm to act as a guide to the Generator improving performance and also supplementing as a weak prior to Zeroth Order Gradient approximation \cite{kariyappa2021maze}, which is employed in most of the Model extraction approaches. \textbf{SHAP} Stands for \textbf{SH}apley \textbf{A}dditive ex\textbf{P}lanations, It calculates feature importance indicating the contributing of sample towards a black box models output Eq. \ref{eq:additivefeature}, This output can be from a regression, classification or any other open ended model. We use introduce a differentiable pipeline that utilizes SHAP values to optimize the generator for custom objectives. Within this pipeline, we optimize the generation for each class by our conditional generator, which enhances the class distribution as evidenced in Fig. \ref{fig:class_dist}. Furthermore, the custom objective enhances sample quality, as demonstrated by Activation Atlases \cite{carter2019activation} in Fig.\ref{fig:shaplucent} is superior to common objectives employed in \cite{truong2021data}, \cite{kariyappa2021maze} in Fig.\ref{fig:nonshaplucent} 

\begin{equation}
f(x) = \mathbb{E}[f(.)] + \sum_{i=1}^{M} \phi_{i} * x'_{i}
\label{eq:additivefeature}
\end{equation}

In this work, our key contributions can be enumerated as below:
\begin{itemize}
    \item We introduce an efficient class-targeting approach for model extraction, significantly enhancing the efficacy of the substitute model across all classes.
    \item We devise a query-efficient feedback mechanism to train a generator which facilitates the pipeline scale to higher dimensional spaces. We demonstrate this through a comparative analysis against prior works, while being the first to extract Video Classification models with an acceptable accuracy and query budget.
    \item Our algorithm's versatility is demonstrated across various settings, including Greybox, BlackBox, BlackBox with soft labels, BlackBox with top-k \& top-1 soft labels, BlackBox with top-k \& top-1 hard labels
\end{itemize}

We also explore the limitations of this approach and provide considerations one should take into account when employing this strategy. To support further research and development in model extraction attacks, we have made the source code\footnote{\href{https://github.com/vidmodex/vidmodex}{\texttt{https://github.com/vidmodex/vidmodex}}} available publicly.


\section{Related Work}\label{sec:related}

We have outlined the motivation for this work in the introduction; this section will review the seminal literature related to each component or domain critical to our study.

\subsection{Model Extraction Attacks}\label{sec:relmodex}
Previous efforts have attempted to propose algorithms for model extraction in Softlabel settings and compute the approximate gradients for backpropagation of objectives \cite{truong2021data}, \cite{miura2021megex}, \cite{beetham2023dualstudent}, \cite{kariyappa2021maze}. Works such as \cite{sanyal2022towards} extensively evaluate pipelines for Hardlabel settings, establishing a precedent for real-world model extraction. Although these approaches utilize similar frameworks with varied mechanisms for training the Generator, they share a common goal: optimizing the divergence between the Victim and Substitute as a discriminator for the Generator. However, these works are also limited in terms of performance due to the query costs required to approximate gradients for a single sample using the Zeroth Order gradient approximation. Efforts have been made to train an efficient Generator using an Evolutionary Algorithm \cite{orekondy2019knockoff}, \cite{barbalau2020black}, \cite{pal2020activethief}, \cite{lin2023quda} have demonstrated significantly lower extraction accuracy compared to the methods previously discussed \cite{oliynyk2023know}. Miscellaneous works like \cite{wang2021zero} focus on generating class-specific samples using minimum decision boundaries, which is superior to other approaches based on sample efficiency to train the Substitute model. However, computing these samples requires a high number of Victim Model queries, making it impractical in real-world scenarios due to the extensive querying required. Inferring from previous work, we understand the influence of samples significantly determines the extraction accuracy and efficiency of the approach. We attempt to address this trade-off by developing an auxiliary objective based on SHAP for the generator that is query-efficient and also improves the fidelity of the generated samples, enabling richer extraction of the Victim model.

\subsection{Interpretable AI for GAN, Model Extraction}\label{sec:relXAI}
Research on utilizing Interpretable AI algorithms for training GANs remains sparse \cite{nagisetty2002xai}, largely because while explanations facilitate human interpretation, they are inferior in information density compared to gradients through the target network and discriminator. On the contrary, these methods have the potential for applications involving black box models, particularly in model extraction \cite{yan2023explanationis}, \cite{wang2022blackdissector}, \cite{oksuz2023autolycus}, \cite{miura2021megex}. While \cite{yan2023explanationis} and \cite{oksuz2023autolycus} simply train victim model to have the same explanation as that of the victim model, this does not direct the model to have better extraction accuracy and \cite{miura2021megex} assumes direct gradients from victim model defeating the purpose. In \cite{wang2022blackdissector} it employs GradCam \cite{selvaraju2017grad} to enhance sample augmentation by focusing on saliency maps from the substitute model to refine the loss function. The approach is constrained because GradCam depends on gradients from the model; Hence, using the substitute model for such operations leads to a noisy and unstable training process. Although the authors demonstrate stability within a confined study using predefined images from a surrogate dataset, the scalability of this approach to diverse real-world objectives remains dubious. While We Iterate on this by computing SHAP\cite{shap2024partitionexplainer} values, which don't require gradient and can therefore be computed directly on the Victim model within a constrained \texttt{max\_evals} budget for each sample. 
While acquiring SHAP values from the Victim model for \texttt{max\_evals} for each sample is costly, we mitigate this by learning to estimate SHAP values within an Energy GAN framework \cite{zhao2016energy}, as deriving SHAP values \cite{jethani2021fastshap} is more feasible than predicting gradients for the victim model.


\subsection{Surrogate Dataset and Settings}\label{sec:relsetting}
In this subsection, we discuss the different settings and the utilization of surrogate datasets in prior research. Numerous studies have employed surrogate datasets \cite{truong2021data}, \cite{sanyal2022towards}, \cite{wang2022blackdissector}, \cite{lin2023quda} each with different assessment on how samples should be selected from the surrogate or proxy dataset. Although these approaches significantly accelerate the extraction process, they require a prior understanding of the data distribution of the victim model, which complicates scalability as \cite{Truong_2021_CVPR_Supp} summarizes the adverse effect of poor surrogate datasets. With the expansion of MLaaS platforms and the increasing number of classes for tasks, several aggregators \cite{google_cloud_vertex_ai_2021}, \cite{amazon_rekognition_video_2024}, \cite{azure_ai_vision_2024}, \cite{azure_custom_vision_2024} now allow entities to deploy their models with the following settings: 1) Top-1 class labels 2) Top-k class labels 3) Top-1 prediction probability 4) Top-k prediction probability. We adopt these settings for our evaluations and also adopt to prediction probability of all classes to ensure comparative analysis with previous research. To offer a framework comparable to other algorithms for utilizing a surrogate dataset, we label the results under the grey box model extraction attack, with further details on the surrogate dataset used for each experiment in Sec.\ref{sec:experiments}. 

\section{Preliminary}\label{sec:prelim}

In this section, we briefly introduce SHAP, an Additive explanation utilizing shapely values, particularly focusing on its application in defining objectives. We employ Partition Explainer\cite{shap2024partitionexplainer}, which recursively computes shapley value through a hierarchy of features; this hierarchy defines feature coalitions and results in the Owen values\cite{lopez2009relationship} from game theory. A detailed view on which is given in Appendix.\ref{A:shapcompute}. Adhering to the fundamental principles of any SHAP explainer, we begin with the additive property presented in Eq. \ref{eq:additivefeature}. In this equation \(f\) represents the target black-box model, \(M\) the size of input space, \(\mathbb{E}[f(.)]\) the expected value of \(f\) over a uniform random distribution and \(\phi\) is the shapley value calculated over for the sample \(x\), expressed as \(\phi (f,x)\). Here \(x_{i}\) represents the \(i^{th}\) feature in \(x\). The relationship between \(x'\) and \(x\) is given by the mapping function \(x = h(x')\) as defined in \cite[Section 2]{lundberg2017unified}, with \(x' \in [0,1]^{M}\) standardised for the algorithms.

For generalizing to the different scenarios outlined in sec.\ref{sec:relsetting}, we define our black-box victim model using Eq.\ref{eq:bbprobdef}. This approach ensures consistent outputs across any top-k prediction setting. Here, \(topk\_probs\) represents the probability values returned for topk predictions, and \(topk\_indicies\) are the indices corresponding to these predictions. The output is a column vector of dimension \([0,1]^{num\_classes}\), depicting a softmax output for a single class prediction scenario, which aligns closely with the intended application within the SHAP framework.

\begin{minipage}{.6\linewidth}
\begin{equation}
f_{sl} = \begin{cases} 
topk\_probs[i] & \text{if } i \in \text{topk\_indices}\vspace{5pt}, \\
\dfrac{1 - sum(topk\_probs)}{num\_classes - k} & \text{otherwise}.
\end{cases}
\label{eq:bbprobdef}
\end{equation}
\end{minipage}
\hspace{-5pt}
\begin{minipage}{.4\linewidth}
\begin{equation}
f_{hl} = 
\begin{cases} 
1 / \text{k} & \text{if } i \in topk\_labels, \\
0. & \text{otherwise }
\end{cases}
\label{eq:bbharddef}
\end{equation}
\end{minipage}

For hard labels, we utilize the definition specified in Eq. \ref{eq:bbharddef}. which assigns a straightforward binary output from the target black box model. We further explore how this method, although it conveys less information compared to the soft label approach, is adequately informative for calculating shapley values.

With the help of the definition Eq.\ref{eq:additivefeature} which is a approximation under the local accuracy property given in \cite[Section 3]{lundberg2017unified} and with choosing either function from Eq.\ref{eq:bbprobdef} or Eq.\ref{eq:bbharddef}, we derive Eq. \ref{eq:classwisederiveA}. We then represent the variables in the equation as vectors, reformulating \(\phi = (\phi_1, \dots, \phi_i, \dots)^\top \) as a column vector and \(x' = (x'_1, \dots, x'_i, \dots)^\top\) as a column vector, regardless of their original shape to obtain Eq. \ref{eq:classwisederiveB}. By applying this framework and forcing on specific class id \(c\), we refine the formula to Eq. \ref{eq:classwisederiveC}.

\begin{subequations}
    \vspace{-8pt}
    \begin{align}
    \begin{split}
        f(x) \ \ \ & = \mathbb{E}[f(.)] + \sum_{i=0}^{M} \phi_{i} * x_{i}
    \label{eq:classwisederiveA}
    \end{split}\\
    \begin{split}
        f(x)\ \ \  & = \mathbb{E}[f(.)] + \phi (f, x)^\top * x'
    \label{eq:classwisederiveB}
    \end{split}\\
    \begin{split}
        f(x|c) & = \mathbb{E}[f(.|c)] + \phi(f(.|c), x) ^\top * x'
    \label{eq:classwisederiveC}
    \end{split}
    \end{align}
\label{eq:classwisederive}
    \vspace{-4pt}
\end{subequations}

Using \ref{eq:classwisederive} we define our objective to improve samples \(x\) to maximize class probability of the targeted model \(f(.|c)\). Hence, we obtain eq. \ref{eq:classwise_obj} as \(\mathbb{E}[f(.|c)]\) is not dependent on \(x\); we further simplify by replacing \(x'\) to reduce the final objective to be linearly proportional to the variable of interest. Hence we use \(j\) which is a column vector with 1's \( j = (1, 1, \dots, 1)^\top\) of size \(M\) as \(x' \in [0,1]^M\) we use the upper bound and lower bound of the objective to be \(0 \leq \phi^\top * x' \leq \phi^\top * j\) or \( 0 \geq \phi^\top * x' \geq \phi^\top * j\) based on the signum of \(\phi^\top * x'\). While the use of \(\phi^\top * j\) increases inaccuracy in objective, it enables us to shift the focus towards the contribution and coalition of each feature and not their magnitude. This was also another decision choice to solve the problem of exploding gradients observed while training. Finally, giving us the objective in Eq.\ref{eq:classobj}

\begin{equation}
    \begin{aligned}
        \underset{x}{\arg\max} \quad f(x|c) & =  \underset{x}{\arg\max} \quad \mathbb{E}[f(.|c)] + \phi (f(.|c), x)^\top * x'\\
        & = \underset{x}{\arg\max} \quad \phi (f(.|c), x)^\top * x'
    \end{aligned}
\label{eq:classwise_obj}
\end{equation}
\begin{equation}
    \begin{aligned}
       ClassObj & = \underset{x}{\arg\max} \quad \phi (f(.|c), x)^\top * j \text{\quad or \quad} ClassObj = \underset{x}{\arg\max} \quad  \sum_{i=1}^{M} \phi_{i} (f(.|c), x) \\
    \end{aligned}
\label{eq:classobj}
\end{equation}

Alongside the previously defined objective, a few crucial parameters are employed, which influence the accuracy of the approximations used in Eq. \ref{eq:additivefeature}. The crucial parameters are: 1) \texttt{max\_evals}: Partition explainer efficiently distributes Shapley value computations across a feature hierarchy rather than exhaustively calculating values for each feature. This approach significantly reduces the inference cost in high-dimensional settings, avoiding \(M!\) inferences and instead requiring only \texttt{max\_evals} calls to the targeted model. 2) \texttt{masker}: The Partition Explainer differs from other explainers by excluding either a single unit or multiple units of features at once, rather than offering granular control over each individual feature. The \texttt{masker} parameter defines this granularity we employ a \(3\times3\) \& \(3\times3\times3\) block of pixel for image \& video models respectively. To obscure these sections, we use Gaussian Blur instead of zero-filling. This technique effectively reduces information without adversely affecting model predictions. Both \texttt{max\_evals} \& \texttt{masker} are highly influential and dependant on the target model's complexity and nuances.

\section{Approach}\label{sec:approach}

\begin{figure}[t!]
    \centering
    \includegraphics[width=\textwidth]{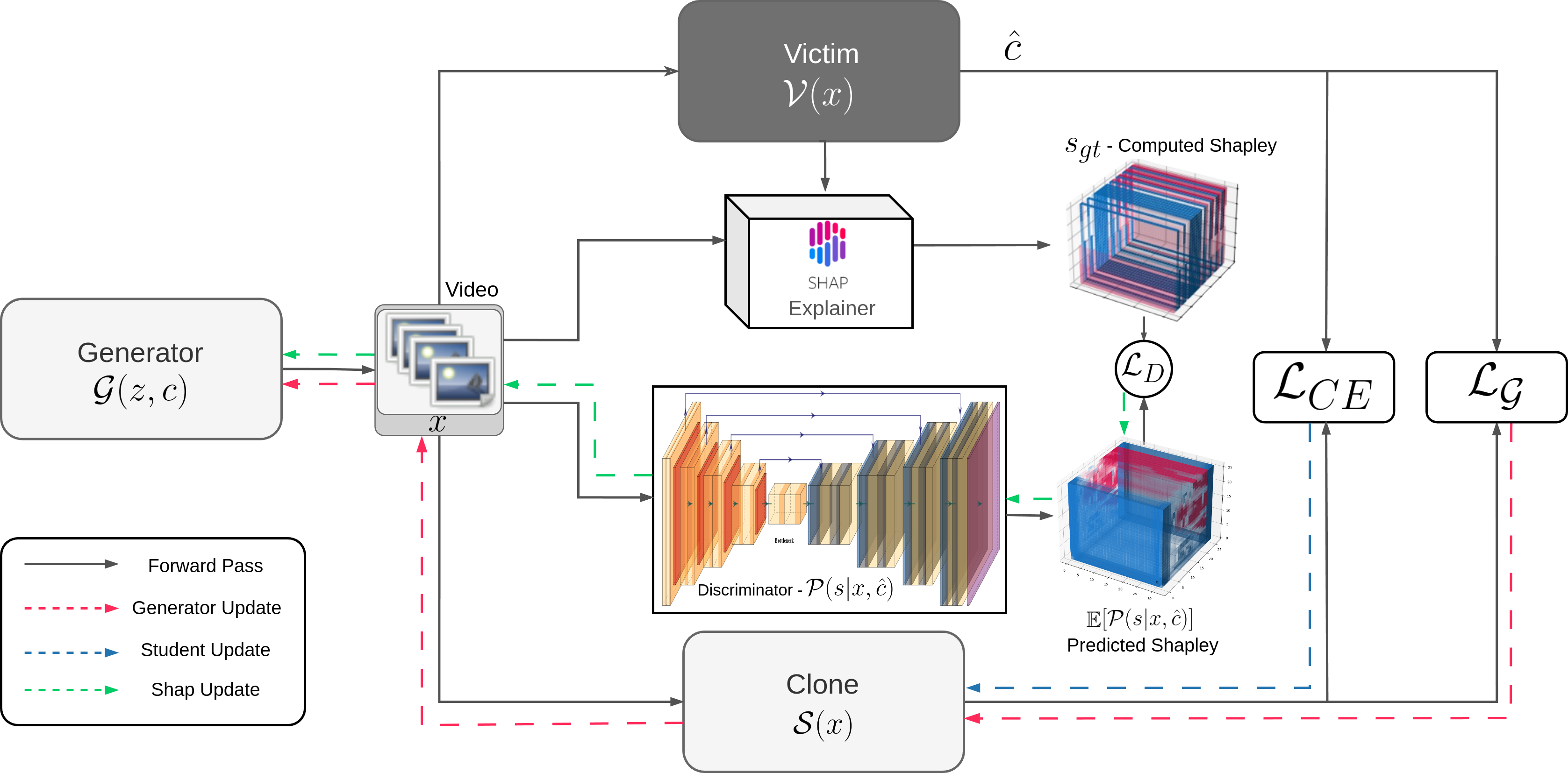}
    \caption{Model extraction diagram with additional objectives and SHAP explainers}
    \label{fig:approach}
    \vspace{-5pt}
\end{figure}



The overall attack setup is well outlined by previous works \cite{truong2021data}, \cite{sanyal2022towards}, with \(\mathcal{V}\) the Victim black box model, \(\mathcal{S}\) a substitute model and A generator \(\mathcal{G}\) which is responsible for crafting input samples. While our objective is to learn \(\mathcal{S}\) that closely mimics the \(\mathcal{V}\). We employ KL divergence\cite{truong2021data} for soft label setting given in Eq.\ref{eq:softlabel_cloneloss}, and employ CrossEntropy Loss\cite{sanyal2022towards} for hard label setting given in Eq.\ref{eq:hardlabel_cloneloss} to optimize \(\mathcal{S}\). To optimize \(\mathcal{G}\), we use an adversarial loss to increase the divergence between Student and victim model\cite{truong2021data, sanyal2022towards} which is given by Eq. \ref{eq:generatorobj} As we use Conditional Generator instead we also specify \(c_{T}\) Target class index to generate samples for a particular class.

\begin{minipage}{.57\linewidth}
\begin{subequations}
    \begin{align}
        \begin{split}
            \mathcal{L}_{sl}(x) = \sum_{i \ \in \ topk\_indices} \mathcal{V}(x|i)  \log \frac{\mathcal{V}(x|i)}{\mathcal{S}(x|i)}
        \label{eq:softlabel_cloneloss}
        \end{split}\\
        \begin{split}
            \mathcal{L}_{hl}(x) =  - \sum_{i \ \in \ topk\_indices} \mathcal{V}(x|i) * \log(\mathcal{S}(x|i))
        \label{eq:hardlabel_cloneloss}
        \end{split}
    \end{align}
\end{subequations}
\end{minipage}
\begin{minipage}{.4\linewidth}
\begin{equation}
    \begin{aligned} 
        z & \sim \mathcal{N} (0, 1)\\
        x & = G(z, c_{T})\\
        \implies  & \underset{\theta_{\mathcal{G}}}{argmax} \quad  \underset{\theta_{\mathcal{S}}}{argmin} \quad \mathcal{L}(x)
    \label{eq:generatorobj}
    \end{aligned}
\end{equation}
\end{minipage}

Alongside this setup, we additionally employ the ClassWise Objective defined in Eq.\ref{eq:classobj}, while \(\phi\) value obtained from the explainer is not differentiable we introduce a estimator \(\mathcal{P}(s|x,c_{T})\) which estimates the SHAP value of the input given the input sample and the targetted class index. The architecture of \(\mathcal{P}\) is a conditional UNet such that the shape of the predicted shap values is sample as the original input. The model \(\mathcal{P}\) predicts a normal distribution of the SHAP value, this format is also helpful to compute the probability value of \(s_{gt}\) in the predicted distribution. We use the \(\mathcal{P}(s_{gt} | x,c)\) as a mask to the objective to reduce error due to divergence between the prediction and the ground truth. We use this modified objective if \(s_{gt}\) is precomputed for the sample else use the initial objective as given in Eq.\ref{eq:Pobjective}. To improve the estimation of the model \(\mathcal{P}\) we use Mean Absolute Error between the sampled shap value from \(\mathcal{P}\) and shap value computed from the explainer given in Eq. \ref{eq:Ploss}. 

\begin{minipage}{0.63\linewidth}
    \begin{equation}
        \begin{aligned}
            ClassObj &= \underset{x}{argmax} \ \ \sum \mathbb{E}[\mathcal{P}(s|x,c)]\\
             & = \underset{x}{argmax} \sum \mathbb{E}[\mathcal{P}(s|x,c)] \odot \mathcal{P}(s_{gt}|x,c) 
        \end{aligned}
        \label{eq:Pobjective}
    \end{equation}
\end{minipage}
\begin{minipage}{0.36\linewidth}
    \begin{equation}
        \begin{aligned}
            \mathcal{L}_{\mathcal{P}}  = & \sum | s_{gt} - \hat{s} | \ , \\
            & \text{where } \hat{s} \sim \mathcal{P}(x,c)
        \end{aligned}
        \label{eq:Ploss}
    \end{equation}
\end{minipage}

\begin{figure}[t!]
    \centering
    \begin{subfigure}[b]{0.19\linewidth}
        \includegraphics[width=\linewidth,clip,trim=3.6cm 1.3cm 3.2cm 1.3cm]{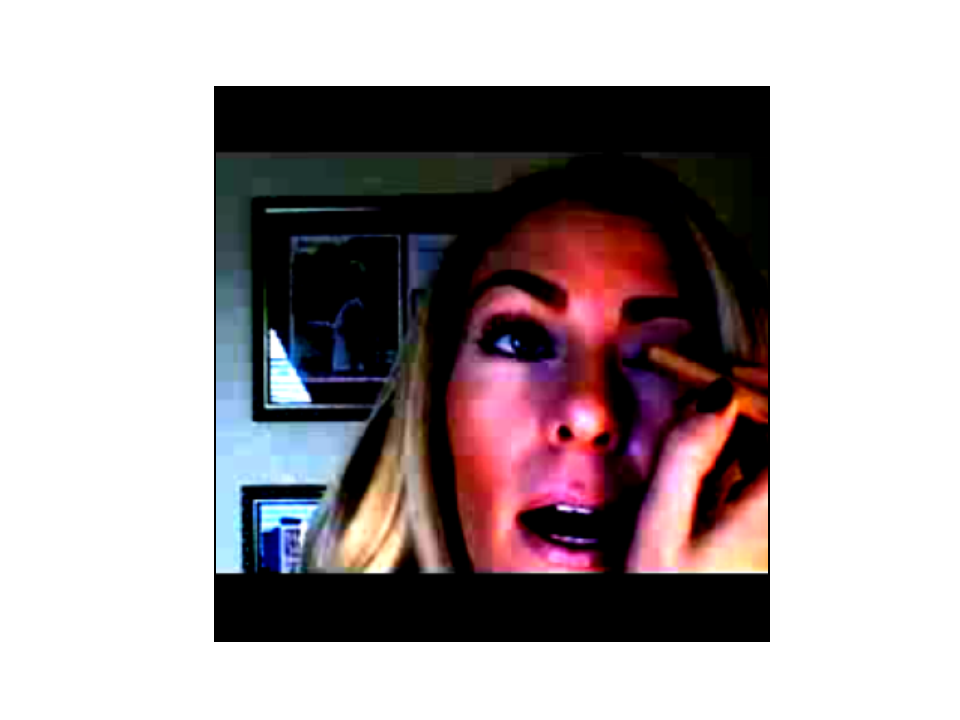}
        \caption{\(x\) - Original input}
        \label{fig:shapOriginalInput}
    \end{subfigure}
    \hfill
    \begin{subfigure}[b]{0.19\linewidth}
        \includegraphics[width=\linewidth,clip,trim=3.6cm 1.3cm 3.2cm 1.3cm]{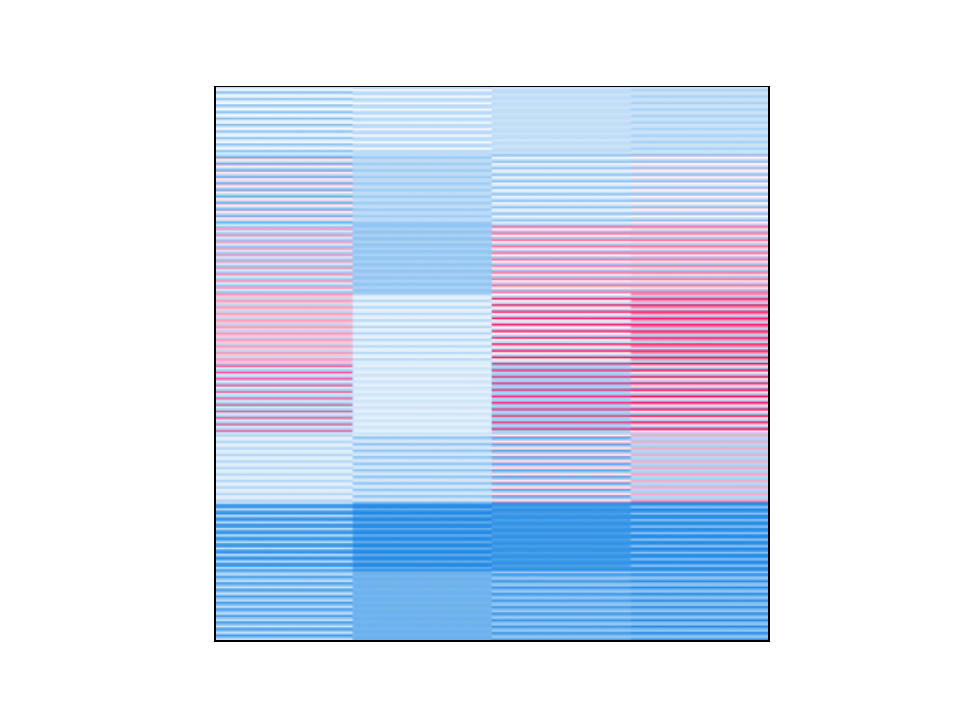}
        \caption{\(s_{gt}\) - SHAP GT}
        \label{fig:shapOriginalShap}
    \end{subfigure}
    \hfill
    \begin{subfigure}[b]{0.19\linewidth}
        \includegraphics[width=\linewidth,clip,trim=3.6cm 1.3cm 3.2cm 1.3cm]{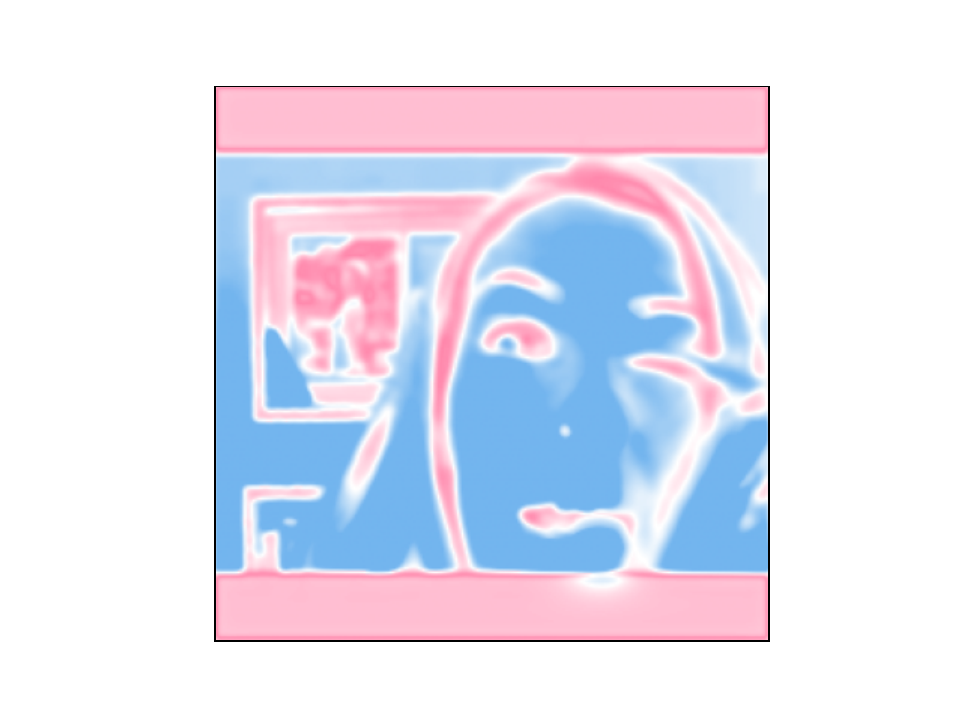}
        \caption{\(\mathbb{E}[\hat{s}]\) - SHAP Pred}
        \label{fig:shapMuShap}
    \end{subfigure}
    \hfill
    \begin{subfigure}[b]{0.19\linewidth}
        \includegraphics[width=\linewidth,clip,trim=3.6cm 1.3cm 3.2cm 1.3cm]{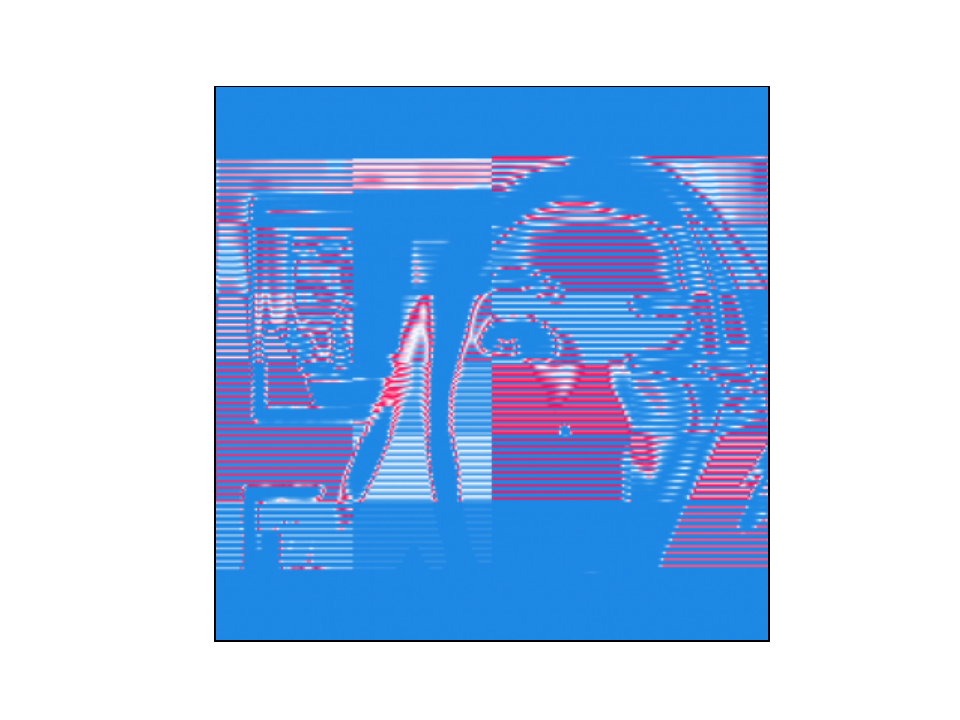}
        \caption{\(\mathcal{P}(s_{gt})\) - P Mask}
        \label{fig:shapProbShap}
    \end{subfigure}
    \hfill
    \begin{subfigure}[b]{0.19\linewidth}
        \includegraphics[width=\linewidth,clip,trim=3.6cm 1.3cm 3.2cm 1.3cm]{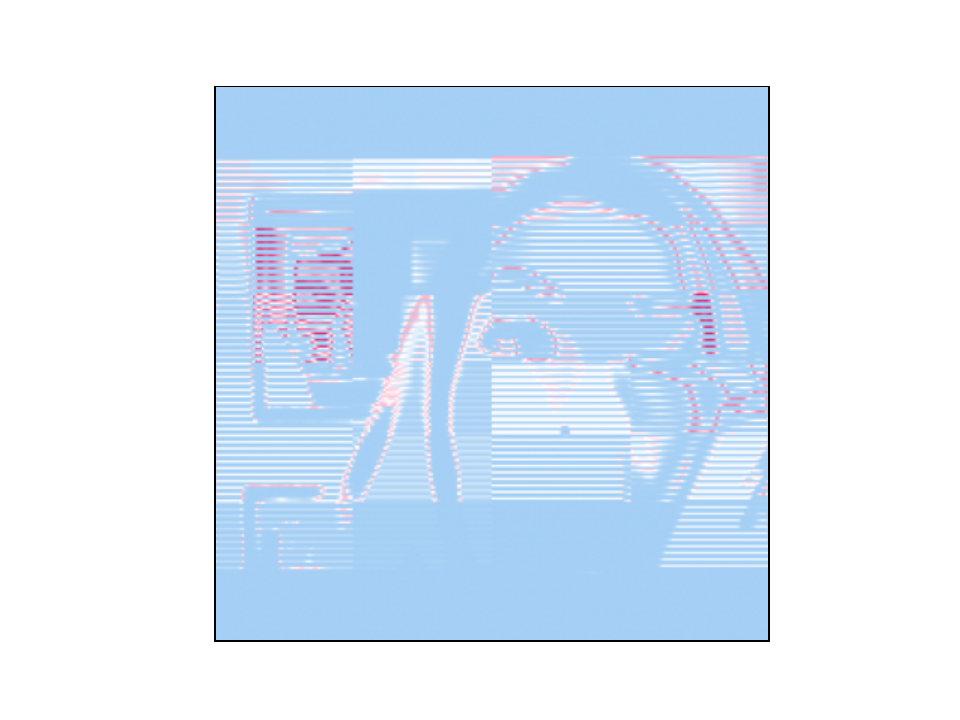}
        \caption{SHAP \textit{Energy}}
        \label{fig:shapEnergyShap}
    \end{subfigure}
    \caption{Shap values and visualization at each stage of the Pipeline}
    \label{fig:shapDiagrams}
\end{figure}

The complete pipeline can be seen in Fig. \ref{fig:approach}, the shap estimator \(\mathcal{P}\) is equivalent to a energy based discriminator in \cite{zhao2016energy} only difference being they are not necessarily in an adversarial setting as \(\mathcal{P}\) is attempting to improve accuracy in estimating the SHAP values for the generated samples, while \(\mathcal{G}\) is being optimized generated samples to increase SHAP values of the samples they generate.  Hence, in the paper, we call \(\mathcal{P}\) and its objective as discriminator in the paper. This objective enables the generator to produce rich samples and improve samples to be balanced across classes. The probabilistic discriminator has an additional mask \(\mathcal{P}(s_{gt}|x,c)\), which ensures no out-of-distribution or noise is introduced while training these components. We normalize the SHAP values to a range of \([-1,1]\) as the magnitude of the shap values varies from \(1 \times 10^{-8} - 1 \times 10^{-11}\) based on datasets and problem settings like image and video models.

Fig. \ref{fig:shapDiagrams} presents a series of visualizations that illustrate the data at each stage within the pipeline. Fig.\ref{fig:shapOriginalInput} displays the initial input to the victim model. This image is a substitute for a generated sample to simplify the interpretation of subsequent images. Fig.\ref{fig:shapOriginalShap} shows the SHAP value computed using the partition explainer. Fig. \ref{fig:shapMuShap} is the expected value \(\mu\) of the discriminator, denoted as \(\mathbb{E}[\mathcal{P}(s|x,c)]\). Fig.\ref{fig:shapProbShap} is the probability mask used to stabilize the initial training phase. It computes the probability that the expected output \(s_{gt}\) matches the predicted distribution, essentially assessing the accuracy of the predictions relative to the ground truth. Finally, Fig.\ref{fig:shapEnergyShap} illustrates the final objective used to train the generator, as specified in Eq. \ref{eq:Pobjective} in energy gan-like architecture.





\section{Experiments}\label{sec:experiments}

This section evaluates our Vidmodex approach under diverse and challenging settings, detailed in sec. \ref{sec:relsetting}, using both image and video models across various datasets such as MNIST\cite{deng2012mnist}, CIFAR10, CIFAR100\cite{krizhevsky2009cifar}, Caltech101\cite{caltec101_li_andreeto_ranzato_perona_2022}, Caltech256\cite{caltech256_griffin_holub_perona_2022} and ImageNet1K\cite{imagenet1k} for images, and UCF11\cite{liu2009ucf11}, UCF101\cite{soomro2012ucf101}, Kinetics 400\cite{kay2017kinetics400}, Kinetics 600\cite{carreira2018shortkin600} and Something-Something v2\cite{goyal2017something} for videos. These tests evaluate over increasing number of classes and complexities; we ensure the evaluation of image model extraction on high-resolution datasets to demonstrate efficiency in large search space. We compare the benchmark primarily across DFME\cite{truong2021data}, DFMS-HL\cite{sanyal2022towards} which we reproduce with our best efforts. We additional utilizes results reported from ZSDB3KD\cite{wang2021zero}, MAZE\cite{kariyappa2021maze}, KnockoffNets\cite{orekondy2019knockoff} and BlackBox Dissector\cite{wang2022blackdissector}. We chose not to replicate results from other studies since our selected methods have already outperformed them in prior works. 
Further, we assess the impact of \texttt{max\_evals} on the extraction process and learning within the discriminator in Sec.\ref{sec:discriminator}, identifying what we consider the best configuration. We conduct an ablation study to explore performance variations across different \texttt{top\_k} settings. Our research primarily focuses on black box model extraction, but we also examine the implications of employing a surrogate dataset (grey box access), discussing its influence later in the paper. Additionally, we present hard label results with analysis over the availability of \texttt{top\_k} labels to validate generalization. Our comprehensive qualitative analysis is detailed in the Appendix. \ref{A:qualitative}, further grounding our findings for the presented empirical evidence.

\subsection*{Experimental setup}\label{sec:expsetup}
DFME and DFMS-HL are integrated as configurable approaches within our pipeline, sharing similar outlines. We also provide scripts to facilitate reproducing these results in the code base. The experiments were conducted on the following hardware setup: 2 nodes of 8 x H100 GPUs(80GB), Intel(R) Xeon(R) Platinum 8480C CPUs (96 cores at 4 GHz Max boost), and 1.8 TB of RAM. We also test all the scripts on a machine with 4 x A100 GPUs (80GB), AMD EPYC 7V13 (64 cores at 4.8 GHz) and 867 GB RAM. While our primary setup is adept, we verify that our scripts run on a modest V100 GPUs(32 GB) system, ensuring reproducibility and facilitating development. Only the experiments involving Kinetics400, Kin600 and Something-Something v2 require the more capable system.

\subsection{Results}

\begin{wraptable}{r}{0.55\linewidth}
\vspace{-10pt}
\resizebox{\linewidth}{!}{
\begin{threeparttable}
    \centering
    \small
    \caption{\centering Comparision of Blackbox Extraction Techniques on Image Models}\label{tab:blackboxmethods}
    \small
    \begin{tabular}{cC{2.3cm}C{0.7cm}C{0.7cm}C{0.7cm}C{0.95cm}}
    \toprule
    Method & Target Dataset / Victim Model & Victim Train Epochs & Victim Acc.\% & Clone Acc.\% & Query Budget \\ \midrule
    \multirow{6}{1.35cm}{\centering DFME\cite{truong2021data}} & MN\(^\ddagger\) / RN-18\(^\dagger\) & 500 & 99.7 & 92.5 & 4M \\ 
    ~ & C10\(^\ddagger\) / RN-18\(^\dagger\) & 1500 & 97.5 & 87.32 & 10M \\ 
    ~ & C100\(^\ddagger\) / RN-34\(^\dagger\) & 3500 & 76.5 & 62.15 & 25M \\ 
    ~ & CT101\(^\ddagger\) / EN-B7\(^\dagger\) & 8000 & 73.2 & 53.56 & 70M \\ 
    ~ & CT256\(^\ddagger\) / EN-B7\(^\dagger\) & 10500 & 77.1 & 32.52 & 100M \\ 
    ~ & IN1K\(^\ddagger\) / EN-B7\(^\dagger\) & 15000 & 67.3 & 13.23 & 120M \\ \cmidrule(lr){1-6}
    \multirow{6}{1.35cm}{\centering DFMS-SL \cite{sanyal2022towards}} & MN\(^\ddagger\) / RN-18\(^\dagger\) & 500 & 99.7 & \textbf{95.1} & 4M \\ 
    ~ & C10\(^\ddagger\) / RN-18\(^\dagger\) & 1500 & 97.5 & 91.22 & 10M \\ 
    ~ & C100\(^\ddagger\) / RN-34\(^\dagger\) & 3500 & 76.5 & 65.04 & 25M \\ 
    ~ & CT101\(^\ddagger\) / EN-B7\(^\dagger\) & 8000 & 73.2 & 56.46 & 70M \\ 
    ~ & CT256\(^\ddagger\) / EN-B7\(^\dagger\) & 10500 & 77.1 & 38.54 & 100M \\ 
    ~ & IN1K\(^\ddagger\) / EN-B7\(^\dagger\) & 15000 & 67.3 & 23.56 & 120M \\ \cmidrule(lr){1-6}
    \multirow{6}{1.35cm}{\centering Vidmodex} & MN\(^\ddagger\) / RN-18\(^\dagger\) & 500 & 99.7 & 94.6 & 4M \\ 
    ~ & C10\(^\ddagger\) / RN-18\(^\dagger\) & 1500 & 97.5 & \textbf{94.9} & 10M \\ 
    ~ & C100\(^\ddagger\) / RN-34\(^\dagger\) & 3500 & 76.5 & \textbf{69.52} & 25M \\ 
    ~ & CT101\(^\ddagger\) / EN-B7\(^\dagger\) & 8000 & 73.2 & \textbf{68.14} & 70M \\ 
    ~ & CT256\(^\ddagger\) / EN-B7\(^\dagger\) & 10500 & 77.1 & \textbf{64.25} & 100M \\ 
    ~ & IN1K\(^\ddagger\) / EN-B7\(^\dagger\) & 15000 & 67.3 & \textbf{48.54} & 120M \\ \cmidrule(lr){1-6}
    \multirow{2}{1.35cm}{\centering ZSDB3KD \cite{wang2021zero}} & MN\(^\ddagger\) / LN-5\(^\dagger\) & - & 99.33 & \textbf{96.54} & 100M \\ 
    ~ & C10\(^\ddagger\) / RN-18\(^\dagger\) & - & 82.5 & 59.46 & 400M \\ \cmidrule(lr){1-6}
    \multirow{2}{1.35cm}{\centering MAZE \cite{kariyappa2021maze}} & C10\(^\ddagger\) / RN-18\(^\dagger\) & - & 92.26 & 45.60 & 30M \\ 
    ~ & C100\(^\ddagger\) / RN-34\(^\dagger\) & - & 82.5 & 37.20 & 80M \\ \cmidrule(lr){1-6}
    \multirow{2}{1.35cm}{\centering KnockOff Nets\cite{orekondy2019knockoff}} & C10\(^\ddagger\) / RN-18\(^\dagger\) & - & 91.56 & 74.44 & 8M \\ 
    ~ & CT256\(^\ddagger\) / RN-34\(^\dagger\) & - & 78.4 & 55.28 & 8M \\ \bottomrule
    \end{tabular}
    \begin{tablenotes}
        \scriptsize 
        \item{\(^\dagger\)Model Architecture\quad RN-18: ResNet18;\quad RN-34: ResNet34;\quad  EN-B7: EfficientNet-B7;\quad LN-5: LeNet-5}
        \item{\(^\ddagger\)Dataset\quad MN: MNIST;\quad  C10: CIFAR10;\quad  C100: CIFAR100; \quad CT101: Caltech101;\quad  CT256: Caltech256;\quad  IN1K: ImageNet1K}
    \end{tablenotes}
\end{threeparttable}
}
\vspace{-30pt}
\end{wraptable}

We present our results for blackbox extraction results in Sec.\ref{sec:blackboxresult}, While further we analyze the influence of top-k on softlabel and hardlabel setting in Sec.\ref{sec:topkresults}. We additionally share the results of Greybox extraction in Sec.\ref{sec:greyboxresults}.

\subsubsection{BlackBox Extraction}\label{sec:blackboxresult}

For the blackbox extraction, our initial investigation centers on SoftLabel Setting with probabilities of all classes from the victim model, in line with previous studies like \cite{wang2021zero}, \cite{kariyappa2021maze}, \cite{orekondy2019knockoff}. As illustrated in Table. \ref{tab:blackboxmethods}, we present the accuracies for these methods as reported in their work and reproduced numbers from \cite{sanyal2022towards} and \cite{truong2021data} alongside our work. To ensure reproducible comparative study, we provide the training epochs required to replicate the victim models, as prior studies often do not offer standardized or pre-trained weights. We train the Target victim architecture from a random initialized state on the target dataset with all configuration details including seeds are available in our code repository. We employ the same architecture for both the clone and the victim model, thereby eliminating any potential bias that might arise from architectural differences.

We also detail the Query Budget, for reported work we either present the reported value or calculate based on the algorithms described, particulary for \cite{wang2021zero}. Our approach demonstrates higher extraction across most tested configurations except in MNIST, where it performs comparably to \cite{sanyal2022towards} and slightly behind \cite{wang2021zero}. Notably, our method is 25\(\times\) efficient than \cite{wang2021zero} based on Query Budget. We employ uniform Query Budget across the methods we reproduce, where we outperform \cite{truong2021data} and \cite{sanyal2022towards} with equivalent budgets. We see a pattern of reducing extraction accuracy with increase in difficulty in the dataset which is correlated to the increase in resolution and increase in number of classes. The increase in difficulty can also be seen with reducing victim accuracy for a higher training epochs. Our approach outperforms \cite{truong2021data} on a average of 16.45\%, with a maximum improvement of 35.31\%. Comparatively, Vidmodex and DFMS-SL Shows a mean improvement of 11.67\% and a maximum improvement of 25.71\%. These statistics underscore the efficacy of our approach across the evaluated datasets.

For video victim models, we employ similar Softlabel setting where we have probability predictions for all classes from the victim model. To facilitate reproduciblity, we have opted to use ViViT-B/16x2 \cite{arnab2021vivit} and Swin-T\cite{liu2022videoswin}, primarily due to their popularity and the ease of use by the providor libraries. We maintain a uniform Query Budget across all three methods and document the training epoch and accuracy for the victim models. Importantly, we do not utilize any pre-trained weights fro these victim models, ensuring a level playing field, as the clone models also lacks access to pre-trained datasets or weights. This allows use to observe a correlation between the victim training epochs and the Query budget while offering a ealistic evaluation of the victim model accuracy, as opposed to the standard practice of using pre-trained weights for video models. As demonstrated in Table. \ref{tab:blackvid}, our method consistently outperforms DFME and DFMS-SL by a considerable margin, with the disparity growing as the complexity of both image and video models increases.


Our Vidmodex method consistently outperforms \cite{truong2021data} and \cite{sanyal2022towards} in video model extraction. Specifically, Vidmodex achieves a mean improvement of 26.11\% and a maximum improvement of 33.36\% over \cite{truong2021data}. In comparison to \cite{sanyal2022towards}, Vidmodex shows a mean improvement of 21.52\% and a maximum improvement of 31.47\% in clone accuracy. Notably, these enhancements are achieved with a Query Budget that is equal to or lower than those utilized in the other two methods.
\subsubsection{Impact of TopK Setting on Soft and Hardlabel extraction. } \label{sec:topkresults}

\begin{wraptable}{l}{0.5\linewidth}
\vspace{-11pt}
\resizebox{\linewidth}{!}{
\begin{threeparttable}
    \small
    \centering
    \caption{\centering Comparision of Blackbox Extraction on Video Models}\label{tab:blackvid}
    \begin{tabular}{cC{2.3cm}C{0.7cm}C{0.7cm}C{0.7cm}C{0.95cm}}
    \toprule
        Method & Target Dataset / Victim Model & Victim Train Epochs & Victim Acc.\% & Clone Acc.\% & Query Budget \\ \midrule
        \multirow{6}{1.35cm}{\centering DFME\cite{truong2021data}} & U11\(^\ddagger\)/VVT\(^\dagger\) & 800 & 84.96 & 55.27 & 70M \\ 
        ~ & U101\(^\ddagger\)/VVT\(^\dagger\) & 2000 & 74.1 & 43.56 & 200M \\ 
        ~ & K400\(^\ddagger\)/SwT\(^\dagger\) & 8000 & 70.8 & 28.49 & 350M \\ 
        ~ & K600\(^\ddagger\)/SwT\(^\dagger\) & 10000 & 68.4 & 18.26 & 420M \\ 
        ~ & SS2\(^\ddagger\)/SwT\(^\dagger\) & 17500 & 61.1 & 11.42 & 500M \\ \cmidrule(lr){1-6}
        \multirow{6}{1.35cm}{\centering DFMS-SL \cite{sanyal2022towards}} & U11\(^\ddagger\)/VVT\(^\dagger\) & 800 & 84.96 & 61.34 & 70M \\ 
        ~ & U101\(^\ddagger\)/VVT\(^\dagger\) & 2000 & 74.1 & 47.53 & 200M \\ 
        ~ & K400\(^\ddagger\)/SwT\(^\dagger\) & 8000 & 70.8 & 34.56 & 350M \\ 
        ~ & K600\(^\ddagger\)/SwT\(^\dagger\) & 10000 & 68.4 & 20.15 & 420M \\ 
        ~ & SS2\(^\ddagger\)/SwT\(^\dagger\) & 17500 & 61.1 & 16.38 & 500M \\ \cmidrule(lr){1-6}
        \multirow{6}{1.35cm}{\centering Vidmodex} & U11\(^\ddagger\)/VVT\(^\dagger\) & 800 & 84.96 & \textbf{72.64} & 50M \\ 
        ~ & U101\(^\ddagger\)/VVT\(^\dagger\) & 2000 & 74.1 & \textbf{68.23} & 200M \\ 
        ~ & K400\(^\ddagger\)/SwT\(^\dagger\) & 8000 & 70.8 & \textbf{57.45} & 350M \\ 
        ~ & K600\(^\ddagger\)/SwT\(^\dagger\) & 10000 & 68.4 & \textbf{51.62} & 420M \\ 
        ~ & SS2\(^\ddagger\)/SwT\(^\dagger\) & 17500 & 61.1 & \textbf{37.63} & 500M \\ \bottomrule
    \end{tabular}
    \begin{tablenotes}
        \scriptsize 
        \item{\(^\dagger\)Model Architecture\quad VVT: ViViT-B/16x2;\quad SwT: Swin-T;}
        \item{\(^\ddagger\)Dataset\quad U11: UCF-11;\quad U101: UCF-101;\quad K400: Kinetics-400;\quad K600: Kinetics-600;\quad SS2: Something-Something-v2; }
    \end{tablenotes}
\end{threeparttable}
}
\end{wraptable}



\begin{figure}[b!]
    \begin{subfigure}[b]{0.265\linewidth}
        \includegraphics[width=\linewidth,clip,trim=0.05cm 0cm 0.4cm 0.7cm]{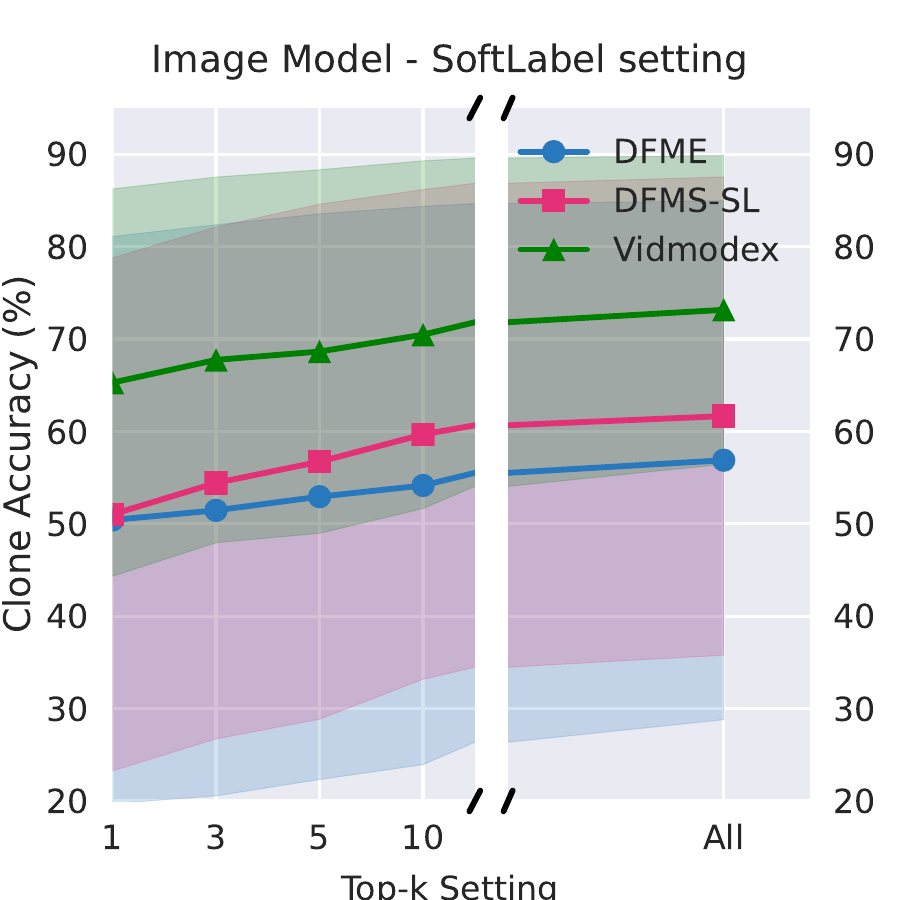}
        \caption{}\label{fig:Ksoftlabelimage}
    \end{subfigure}
    \begin{subfigure}[b]{0.255\linewidth}
        \includegraphics[width=\linewidth,clip,trim=0.65cm 0cm 0.4cm 0.7cm]{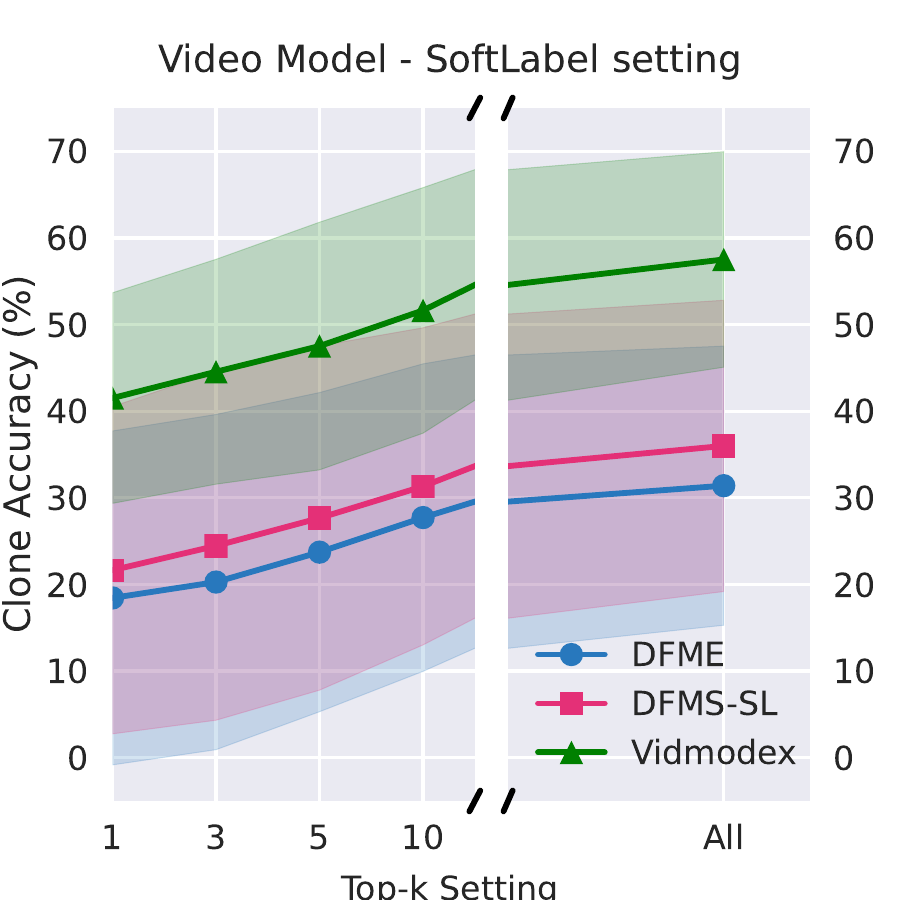}
        \caption{}\label{fig:Ksoftlabelvideo}
    \end{subfigure}
    \begin{subfigure}[b]{0.225\linewidth}
        \includegraphics[width=\linewidth,clip,trim=0.65cm 0cm 1.6cm 0.7cm]{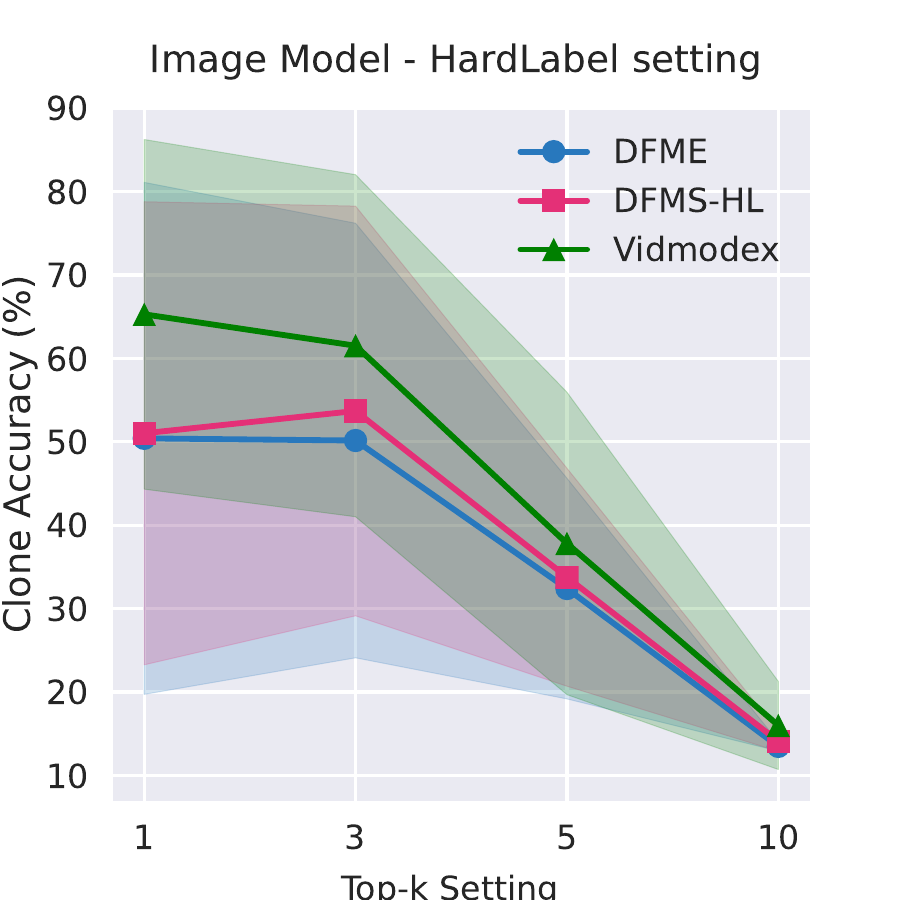}
        \caption{}\label{fig:Khardlabelimage}
    \end{subfigure}
    \begin{subfigure}[b]{0.225\linewidth}
        \includegraphics[width=\linewidth,clip,trim=0.65cm 0cm 1.6cm 0.7cm]{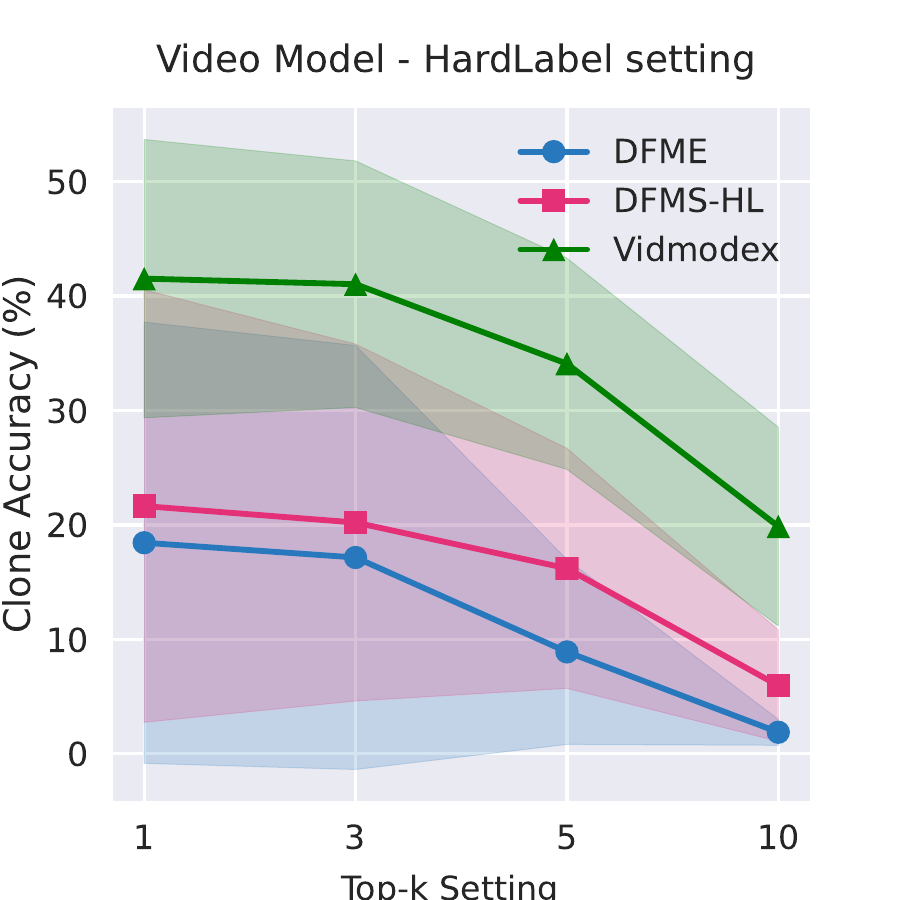}
        \caption{}\label{fig:Khardlabelvideo}
    \end{subfigure}
    \caption{\centering Plots of the extraction accuracy across different K, for both Softlabel and Hardlabel setting}
    \label{fig:topkplots}
\end{figure}

We explore the scenario where top-k labels are available for model extraction, illustrating the real-world applicability of our pipeline. This analysis adheres to the definition provided in Eq.\ref{eq:bbprobdef} for softlabel and Eq.\ref{eq:bbharddef} for hardlabel, ensuring consistency across all scenarios. Our approach does not incorporate any specific methodology for handling top-k labels beyond these definitions. This analysis aims to demonstrate that the computation of the SHAP values and the introduction of the New Shap-based objective do not negatively impact performance when fewer labels are returned. As shown in Fig.\ref{fig:topkplots}, we plot the mean clone accuracy for each value of K, with regions defined by the standard deviations of these points for each K. For Softlabel extraction, we report results for image and video models across K \(\in {1,3,5,10,\text{ALL}}\), While for Hardlabel extraction, we report K \(\in {1,3,5,10}\); the 'All' category is omitted as it would equate to no information for hardlabel; hence the focus is limited to these K values. Notably, for datasets like MNIST, CIFAR10 and UCF11, we do not present numbers for K=10 in hardlabel as the total number of classes is 10 or 11, making hardlabels redundant in such scenarios. From Fig.\ref{fig:Ksoftlabelimage} for image models, there is an observed upward trend in extraction accuracy as richer information is introduced with each increasing label. A similar trend is observed for video models in Fig.\ref{fig:Ksoftlabelvideo}. Conversely, Fig.\ref{fig:Khardlabelimage} and Fig.\ref{fig:Khardlabelvideo} show a reverse trend for hardlabel extraction; accuracy decreases with increasing top-k as information diminishes with each additional label in hardlabel settings. These trends can be directly correlated with the entropy of the victim model in each scenario. Detailed configurations of these experiments are available in the Appendix: Table. \ref{tab:hardlabeltopkImage} details the hard label extraction for image models, Table. \ref{tab:hardlabeltopkVideo} for video models, Table. \ref{tab:softlabelImage} for softlabel extraction of image models, and Table. \ref{tab:softlabeltopkVideo} for softlabel extraction of video models across various K values.

\subsubsection{Grey Box extraction}\label{sec:greyboxresults}

\begin{wrapfigure}{l}{0.7\linewidth}
    \centering\captionsetup[subfigure]{justification=centering}
    \begin{minipage}{0.505\linewidth}
        \includegraphics[width=\linewidth,clip,trim=0.3cm 0.45cm 0.5cm 0.42cm]{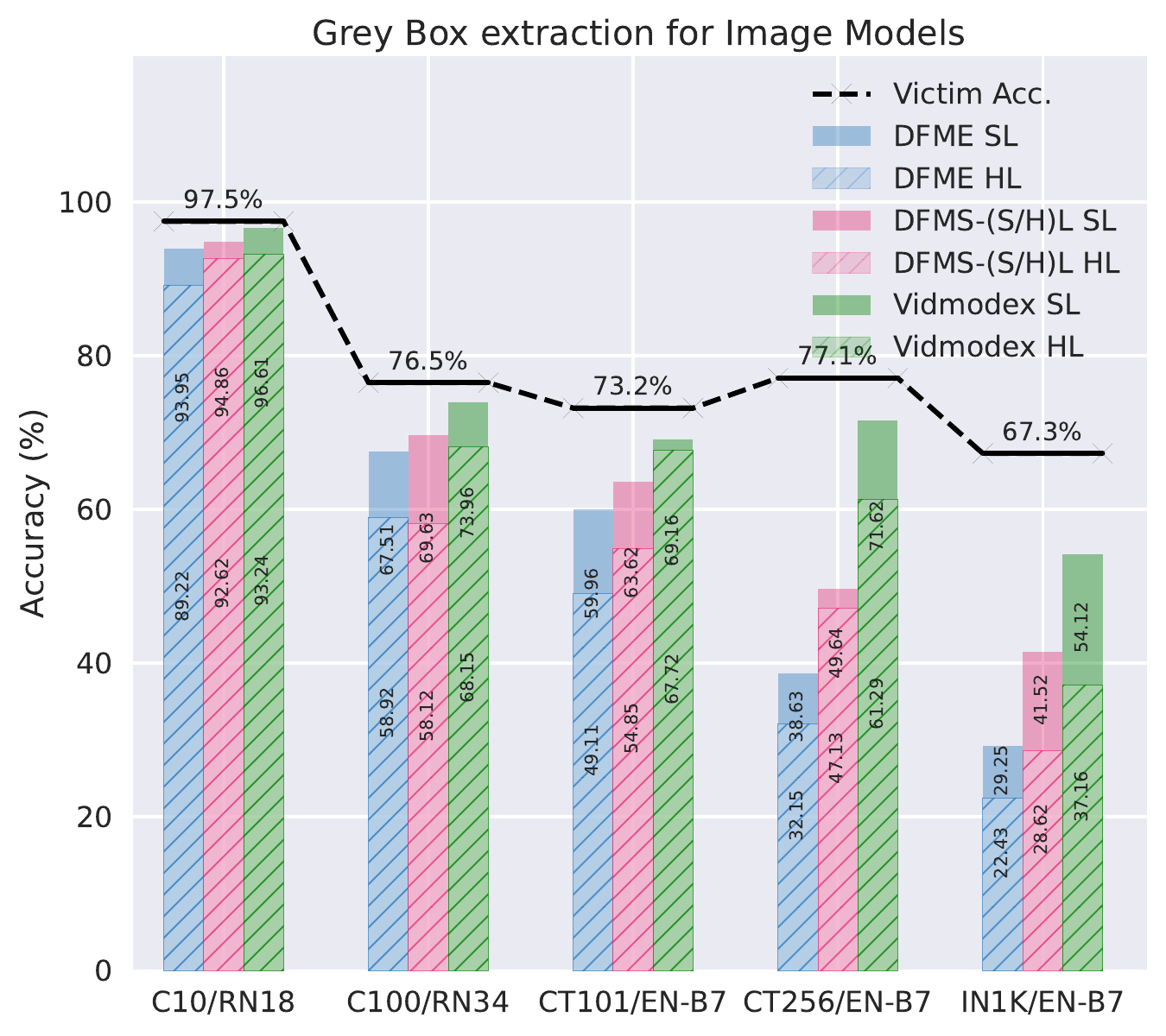}
        \subcaption{\centering Image Model}\label{fig:greyimageplot}
    \end{minipage}
    \begin{minipage}{0.485\linewidth}
        \includegraphics[width=\linewidth,clip,trim=1.25cm 0.45cm 0.5cm 0.42cm]{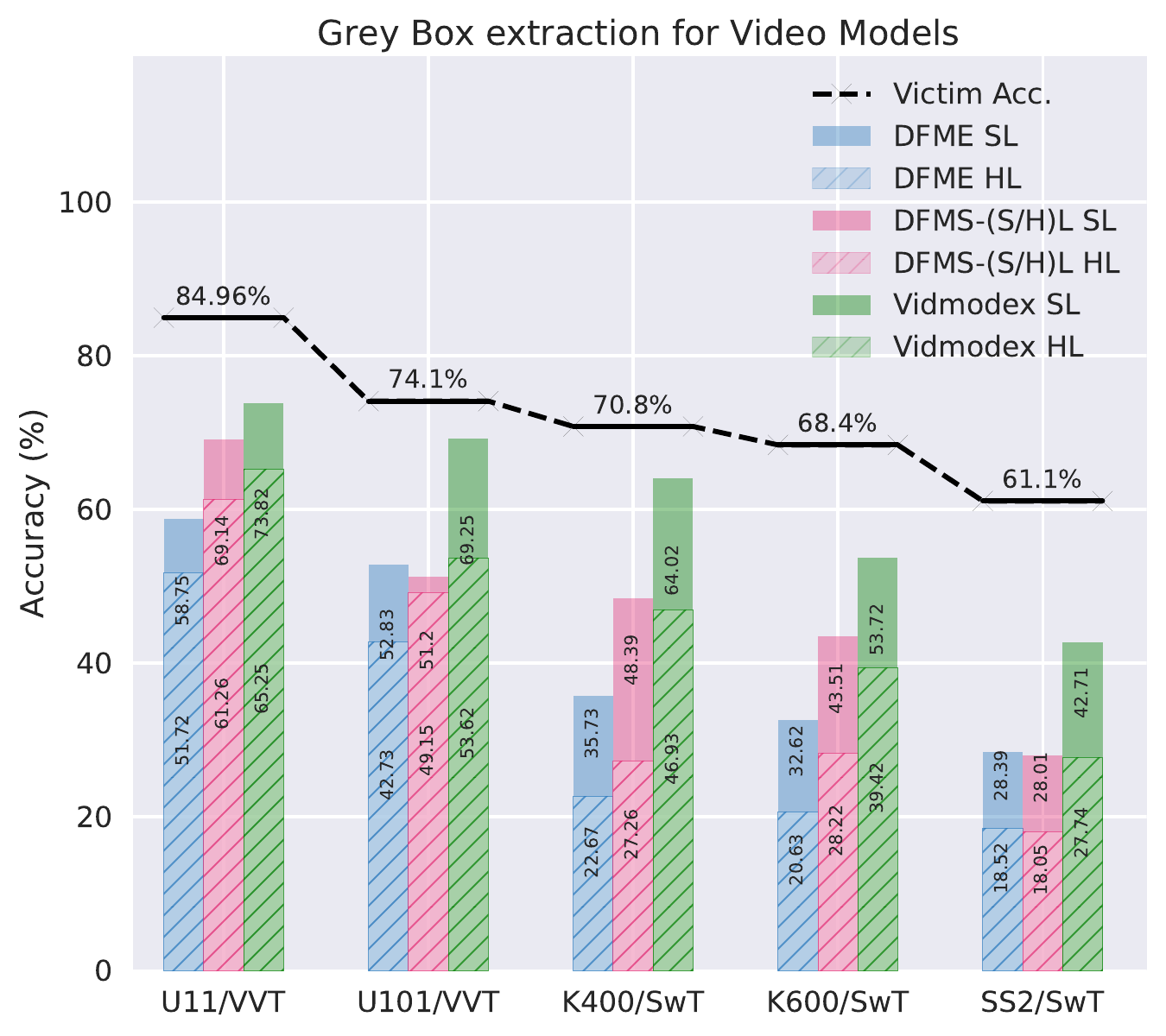}
        \subcaption{\centering Video Model}\label{fig:greyvideoeplot}
    \end{minipage}
    \caption{\centering Comparision of GreyBox extraction methods}\label{fig:greyboxplot}
    \vspace{-5pt}
\end{wrapfigure}

We also evaluate the efficacy of our approach using a surrogate dataset. Although enhancing grey box accuracy is not our primary focus, these experiments serve to confirm that our SHAP-based objective does not adversely affect the generator's learning process when paired with a proxy or surrogate dataset. Instead of delving into the methodology for selecting an appropriate surrogate dataset, we utilize portions of established datasets. Specifically, we incorporate ImageNet-22KK\cite{russakovsky2015imagenet22k} for image models and Kinetics-700\cite{carreira2019shortkin700} and CHARADES\cite{sigurdsson2016hollywoodcharades} for video model extractions. These datasets are shuffled and employed without targeting any specific subclasses.

The experimental details and configurations are outlined in Table. \ref{tab:greyboxImage} for image models and Table.\ref{tab:greyboxVideo} for video models, with results visualized in Fig.\ref{fig:greyboxplot}. Our analysis includes only three methods: \cite{truong2021data}, \cite{sanyal2022towards}, and our own approach, examining both SoftLabel and HardLabel settings. As expected, extraction accuracy declines with increased difficulty, utilizing all labels for SoftLabel settings and only top-1 labels for HardLabel settings.

Despite this complexity, our method demonstrates notable robustness and effectiveness, particularly in the SoftLabel and image model contexts, where it shows a mean improvement of approximately 15.23\% over DFME and 9.24\% over DFMS-SL, with peaks of 32.99\% and 21.98\%, respectively. For the HardLabel setting in image models, our approach yields an average improvement of 15.15\% over DFME and 9.24\% over DFMS-HL, reaching up to 29.14\% and 14.16\%, respectively. In video model extractions under SoftLabel settings, we observe average enhancements of 19.04\% over DFME and 12.65\% over DFMS-SL, with maximum gains of 28.29\% and 18.05\%, respectively. The HardLabel setting shows our method outperforming DFME by 15.34\% and DFMS-HL by 9.80\% on average, with maximum improvements of 24.26\% and 19.67\%, respectively. These findings underscore the robustness and efficacy of our extraction pipeline.

\subsection{Ablation study}
\subsubsection{Discriminator Learning}\label{sec:discriminator}

In this section, we examine the impact of the \texttt{max\_eval} parameter on SHAP value computations, crucial for training the discriminator \(\mathcal{P}\) with the objective defined in Eq. \ref{eq:Ploss}. A higher \texttt{max\_eval} value results in more fine-grained SHAP values for each feature, thereby improving local accuracy as outlined in Eq.\ref{eq:additivefeature} and discussed in \cite{lundberg2017unified}. To enhance the approximation quality of \(\mathcal{P}\), we initially set \texttt{max\_eval} to a high value. However, a high \texttt{max\_eval} also increases the number of victim model queries per sample. To manage this, we progressively reduce \texttt{max\_eval} throughout the training process, similar to learning rate decay techniques \cite{smith2019super}. This reduction strategy halves \texttt{max\_eval} progressively until reaching a minimal threshold, beyond which the discriminator is no longer trained. This eliminates the need for further SHAP computations and reduces queries to the victim model. Additionally, we do not employ masking for SHAP optimization as specified in Eq.\ref{eq:Pobjective}.


\begin{wrapfigure}{r}{0.6\linewidth}
    \includegraphics[width=\linewidth, clip, trim=0.4cm 0.3cm 0.6cm 0.4cm]{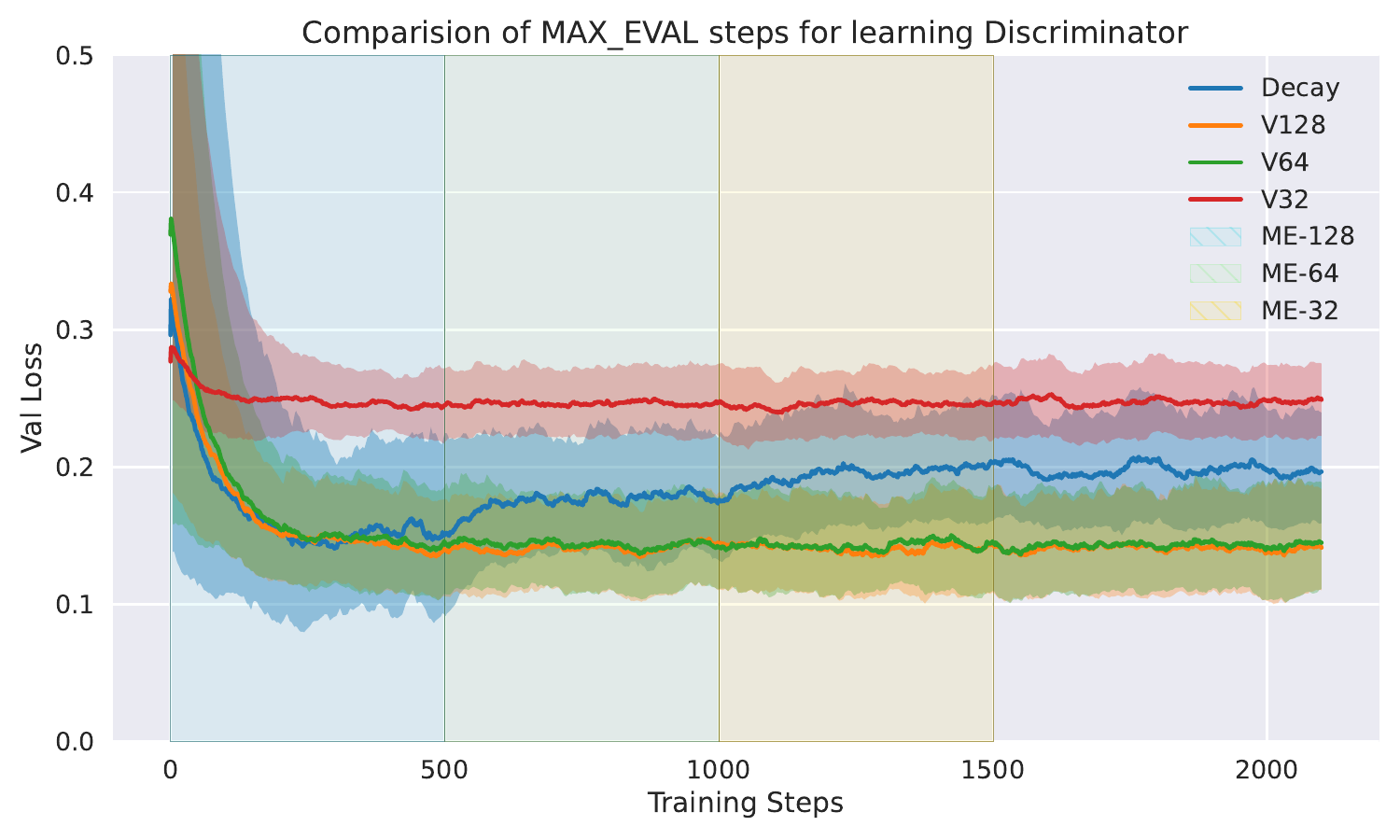}
    \caption{\centering Variance in training Discriminator based on \texttt{max\_evals}}\label{fig:maxevalplot}
\end{wrapfigure}

We explore the stability of our approach using a dynamic \texttt{max\_eval} adjustment mechanism, which can significantly alter the training objective. We conduct an extraction on the CIFAR100 dataset using a ResNet-18 model. The outcome of this experiment is visualized in Fig.\ref{fig:maxevalplot}, where we evaluate the discriminator \(\mathcal{P}\) against the SHAP ground truth, computed with \texttt{max\_eval} set at 1024. For training purposes, we use the smaller values of \texttt{max\_eval} specifically {32, 64, 128}. To establish benchmarks for this configuration, we initially train the discriminator solely to minimize Eq.\ref{eq:Ploss} using fixed \texttt{max\_eval} values of \(\in {32, 64, 128}\). Conversely, for the scheduled decay approach, we implement a descending \texttt{max\_eval} sequence {128, 64, 32} over respective intervals: [(0, 500), (500, 1000), (1000, 1500)]. Our results indicate that while the hybrid decay strategy yields slightly inferior outcomes compared to constant high values of 128 and 64, it significantly outperforms the lowest fixed setting of 32. Moreover, the variance in validation loss under the hybrid approach is lower than 32 and 64 settings, making the approach viable to maintain efficiency while providing substantial feedback for optimizing the pipeline.

\section{Conclusion}\label{sec:conclusion}

In this study, we aimed to enhance the DataFree model extraction framework by integrating Explainable AI algorithms as an auxiliary objective alongside existing methodologies. We rigorously tested our approach in real-world scenarios, encompassing both hard and soft label settings across various top-k outputs, aligning with the constraints typical of contemporary MLaaS offerings. Our research extends the scope of model extraction attacks to video classification models, where we observed significant improvements over previous methods. Both quantitative and qualitative analyses were conducted to assess the impact of SHAP values, underscoring notable enhancements in model extraction. We also detailed the implementation of our pipeline and explored the influence of additional hyperparameters to facilitate reproducibility and further development. Although our approach is broadly applicable to any target model task—including audio, text, and tabular data—this paper focuses on a constrained study to substantiate our claims. Future work could explore the development of generalized extraction techniques for even larger models with billions of parameters, aiming to achieve this at a reasonable cost. While our work primarily details the attack on such models, our overarching goal is to enrich the community's awareness of the substantial MLaaS industry. We believe it is of utmost importance to understand the potential risks involved.



\newpage

\bibliography{references}
\newpage 

\appendix

\section{Appendix / supplemental material}

\subsection{Derivation of Hierarchical Shapley Values from Owen Values}\label{A:shapcompute}

In this section we focus on the Derivation of the Hierarchical Shapley Values, This is implemented as the Partition Explainer. And is defined using the principals of the Owen Value decomposition \cite{lopez2009relationship}. To do so we start with the following Definition \ref{def:additivedefref} and the Theorem \ref{thrm:shapley} from \cite{lundberg2017unified}.
\begin{definition}[Additive feature attribution methods] have an explanation model that is a linear function of binary variables. where \(z^\prime \in \{0, 1\}^{M}\), \(M\) is the number of simplified input features, and \(\phi_{i} \in \mathcal{R}\)
    \label{def:additivedefref}
\end{definition}
\begin{property}[Local Accuracy]
\begin{equation}
    f(x) = g(x') = \phi_0 + \sum_i \phi_i x'_i
\label{eq:localaccuracyproperty}
\end{equation}
The explanation model \(g(x')\) matches the original model \(f(x)\) when \(x = h_x(x')\). 
\end{property}
This is desirable, when approximating the original model \(f\) for a specific input x, local accuracy requires the explanation model to at least match the output of \(f\) for the simplified input \(x'\) (which corresponds to the original input \(x\)).

\begin{property}[Missingness]
\[x'_i=0 \Rightarrow \phi_i = 0\]
Missingness constrains features where \(x'_i=0\) to have no attributed impact.
\end{property}
The second property is missingness. If the simplified inputs represent feature presence, then missingness requires features missing in the original input to have no impact. All of the algorithms obey missingness property.

\begin{property}[Consistency] Let \(f_x(z') = f(h_x(z'))\) and \(z' \setminus i\) denote setting \(z'_i = 0\). For any two models f and f', if
\begin{equation}
f'_x(z') - f'_x(z' \setminus i) \geq f_x(z') - f_x(z' \setminus i)
\label{eq:consistencyproperty}
\end{equation}
for all inputus \(z' \in \{0,1\}^{M}\), then \(\phi_i(f',x) \geq \phi_i(f,x)\).
\end{property}
The third property states that if a model changes so that some simplified input's contribution increases or stays the same regardless of the other inputs, that input's attribution should not decrease.

\begin{theorem} Only one possible explanation model \(g\) follows Definition 1 and satisfies Properties 1, 2 and 3:

\begin{equation}
\phi_i(f, x) = \sum_{z' \subseteq x'} \frac{|z'|!(M - |z'| - 1)!}{M!} \left [f_x(z') - f_x(z' \setminus i) \right] 
\label{eq:shaporiginal}
\end{equation}

Where \(|z'|\) is the number of non-zero entries in \(z'\), and \(z' \subseteq x'\) represents all \(z'\) vectors where the non-zero entries are a subset of the non-zero entries in \(x'\). 
\label{thrm:shapley}
\end{theorem}

Theorem \ref{thrm:shapley} satisfies Property 1, Property 2 and Property 3. Building on the original definition and Theorem \ref{thrm:shapley}, we introduce another property.

\begin{property}[Group Homogeneity] Let \(G_k \subseteq M \) and \(i, j \in G_k\) are in the same group.
\begin{equation}
\mathbb{E}[|\phi_i(f, x) - \phi_j(f, x)|] \leq \varepsilon(|G_k|)
\label{eq:grouphomoproperty}
\end{equation}
where \(\varepsilon(|G_k|)\) is a small positive constant that decreases as \(|G_k|\) decreases.        
\end{property}
This property is crucial for approximation of the SHAPley value for large feature space. This states that for a subset of features in a defined Group, have a close shapley values and the difference between there true values has a upper bound of \(\varepsilon\) which is dependent on the size of the group. This allows us to select a partition of features and approximate the \(\phi\) value to average \(\phi\) value of the group.  

\begin{theorem}[Hierarchical Approximation of SHAP Values]
Let \(f\) be a model with input feature set \(M\), partitioned into \(k\) disjoint groups \(G_1, G_2, \ldots, G_k\), such that \(M = \bigcup_{i=1}^k G_i\) and \(G_i \cap G_j = \emptyset\) for \(i \neq j\). Let \(\phi_i(f, x)\) denote the SHAP value for feature \(i\) given input \(x\).

Under the following properties:
\begin{enumerate}
    \item \textbf{(Local Accuracy)} \(f(x) = g(x') = \phi_0 + \sum_i \phi_i x'_i\)
    \item \textbf{(Missingness)} \(x'_i = 0 \Rightarrow \phi_i = 0\)
    \item \textbf{(Consistency)} If \(f'_x(T \cup \{i\}) - f'_x(T) \geq f_x(T \cup \{i\}) - f_x(T)\) for all \(T \subseteq z'\), where \(z'\in \{0,1\}^M\), then \(\phi_i(f', x) \geq \phi_i(f, x)\)
    \item \textbf{(Group Homogeneity)} For any two features \(i, j\) in the same group \(G_k\), \(\mathbb{E}[|\phi_i(f, x) - \phi_j(f, x)|] \leq \varepsilon(|G_k|)\), where \(\varepsilon(|G_k|)\) is a small positive constant that decreases as \(|G_k|\) decreases
\end{enumerate}

The hierarchical approximation of SHAP values for feature \(i\) in group \(G_k\) is given by:

\begin{equation}
    \phi_i(f, x) \approx \frac{1}{|G_k|} \sum_t \phi_k^t(f)
\label{eq:groupdistshaptheorem}
\end{equation}

where \(\phi_k^t(f)\) is the Owen value for group \(G_k\):

\begin{dmath}
\phi_k^t(f) = \sum_{t: G_k \subseteq T \subseteq M \setminus \{i\}} \frac{|T \cap G_k|!(|G_k|-|T \cap G_k|-1)!}{|G_k|!} \cdot \frac{|T|!(|M|-|T|-1)!}{|M|!} \cdot [f(T \cup \{i\}) - f(T)]
\end{dmath}

This approximation improves as \(\varepsilon(|G_k|)\) decreases with smaller group sizes. However, the optimal group size is determined by balancing approximation accuracy with computational efficiency, constrained by a maximum number of allowed function evaluations \(\text{max\_eval}\).
\end{theorem}

\begin{proof}
The standard SHAP formula for feature i from Theorem \ref{thrm:shapley}:
    \[\phi_i(f, x) = \sum_{z' \subseteq x'} \frac{|z'|!(M - |z'| - 1)!}{M!} [f_x(z') - f_x(z' \setminus i)]\]

Given the partition of \(M\) into groups \(G_1, G_2, \ldots, G_k\), we can rewrite this as:

\begin{equation}
\begin{split}
\phi_i(f, x) = \sum_k \sum_{T_k: G_k \subseteq T \subseteq M \setminus \{i\}} \frac{|T_k \cap G_k|!(|G_k|-|T_k \cap G_k|-1)!}{|G_k|!} & \cdot \frac{|T|!(|M|-|T|-1)!}{|M|!} \\
& \cdot [f(T \cup \{i\}) - f(T)]\\
\label{eq:groupparitionshapproof}
\end{split}
\end{equation}

where \(T_k\) represents subsets containing group \(G_k\).

By Property 4 (Group Homogeneity), for any two features \(i, j\) in \(G_k\):

\[\mathbb{E}[|\phi_i(f, x) - \phi_j(f, x)|] \leq \varepsilon(|G_k|)\]

This allows for the approximation of the SHAP value of any feature in \(G_k\) by the average SHAP value of the group:

\[\phi_i(f, x) \approx \frac{1}{|G_k|} \sum_{i \in G_k} \phi_i(f, x)\]

The group's contribution can be expressed using Owen values:

\begin{equation}
\frac{1}{|G_k|} \sum_{i \in G_k} \phi_i(f, x) = \frac{1}{|G_k|} \sum_t \phi_k^t(f)
\end{equation}

where \(\phi_k^t(f)\) is the Owen value for group \(G_k\) as defined in the theorem statement.

Properties 1-3 are preserved in this approximation:
\begin{enumerate}
    \item Local Accuracy: The sum of all group contributions equals \(f(x) - \mathbb{E}[f(x)]\), preserving local accuracy.
    \item Missingness: If \(x'_i = 0\), the entire group \(G_k\) containing \(i\) will have zero contribution, satisfying missingness.
    \item Consistency: Group average contributions maintain consistency if individual feature contributions are consistent.
\end{enumerate}

As \(\varepsilon(|G_k|)\) approaches 0 with decreasing group size:

\[\lim_{|G_k| \to 1} \varepsilon(|G_k|) = 0\]

implying that all features within a group have identical SHAP values in the limit, and the approximation becomes exact as group sizes approach 1.

The hierarchical approximation reduces the computational complexity from \(O(2^M)\) to \(O(2^k)\), where \(k\) is the number of groups and \(k \ll M\). To balance accuracy and computational efficiency, an optimal group size is determined by:

\[\text{optimal\_group\_size} = \arg \min_{|G_k|} \varepsilon(|G_k|) \text{ subject to } 2^k \leq \text{max\_eval}\]

where \(\text{max\_eval}\) is the maximum number of allowed function evaluations.

\end{proof}



\section{Additional Comparative Results}

\subsection{Soft Label}\label{A:softlabel}


For a more detailed analysis, we present experimental results on the variance of top-k values for SoftLabel Black Box extraction. The results are summarized in Figures  \ref{fig:Ksoftlabelimage} and \ref{fig:Ksoftlabelvideo}, with further details provided in Section. \ref{sec:topkresults}. We have comprehensively analyzed a sufficient range of top-k values, specifically 1, 3, 5, 10, and All. These results are detailed in Table. \ref{tab:softlabelImage} for image models, which includes information about the target victim model, the clone models, victim and extraction accuracy, and the query budgets for each experiment. The dataset on which the victim model was trained, including the number of training epochs, is also listed. Similarly, results for the video model are presented in Table. \ref{tab:softlabeltopkVideo}, including all the aforementioned metrics.

\begin{minipage}{0.495\textwidth}

\resizebox{\linewidth}{!}{
\begin{threeparttable}
    \centering
    \small
    \caption{Comparision of SoftLabel Blackbox Extraction on Image Models across TopK}\label{tab:softlabelImage}
    \begin{tabular}{cC{2.3cm}C{0.7cm}C{0.7cm}C{0.7cm}C{0.7cm}C{0.7cm}C{0.7cm}C{0.7cm}C{0.95cm}}
    \toprule
        Method & Target Dataset / Victim Model & Victim Train Epochs & Victim Acc.  & \multicolumn{5}{C{5cm}}{Clone Accuracy with given top-k classes in \%}  & Query Budget \\ \cmidrule(lr){5-9}
        & & & &1 & 3 & 5 & 10 & All & \\ \midrule
        \multirow{6}{*}{DFME\cite{truong2021data}} & MN\(^\ddagger\) / RN18\(^\dagger\) & 500 & 99.7 & 90.1 & 91.8 & 92.2 & 92.5 & 92.5 & 4M \\ 
        ~ & C10\(^\ddagger\)  / RN18\(^\dagger\) & 1500 & 97.5 & 85.22 & 85.83 & 86.21 & 87.32 & 87.32 & 10M \\ 
        ~ & C100\(^\ddagger\)  / RN34\(^\dagger\) & 3500 & 76.5 & 52.67 & 54.63 & 58.25 & 59.74 & 62.15 & 25M \\ 
        ~ & CT101\(^\ddagger\)  / EN-B7\(^\dagger\) & 8000 & 73.2 & 44.96 & 45.62 & 48.21 & 49.61 & 53.56 & 70M \\
        ~ & CT256\(^\ddagger\)  / EN-B7\(^\dagger\) & 10500 & 77.1 & 26.71 & 27.3 & 27.5 & 28.3 & 32.52 & 100M \\ 
        ~ & IN1K\(^\ddagger\)  / EN-B7\(^\dagger\) & 15000 & 67.3 & 2.8 & 3.56 & 5.25 & 7.4 & 13.23 & 120M \\ \cmidrule(lr){1-10}
        \multirow{6}{1.35cm}{\centering DFMS-SL \cite{sanyal2022towards}} & MN\(^\ddagger\)  / RN18\(^\dagger\) & 500 & 99.7 & 86.6 & 89.34 & 91.5 & 95.1 & 95.1 & 4M \\ 
        ~ & C10\(^\ddagger\)  / RN18\(^\dagger\) & 1500 & 97.5 & 81.29 & 86.46 & 89.92 & 91.24 & 91.24 & 10M \\ 
        ~ & C100\(^\ddagger\)  / RN34\(^\dagger\) & 3500 & 76.5 & 53.62 & 57.93 & 59.61 & 60.64 & 65.04 & 25M \\ 
        ~ & CT101\(^\ddagger\)  / EN-B7\(^\dagger\) & 8000 & 73.2 & 46.7 & 48.52 & 49.8 & 51.53 & 56.46 & 70M \\ 
        ~ & CT256\(^\ddagger\)  / EN-B7\(^\dagger\) & 10500 & 77.1 & 32.35 & 32.84 & 35.85 & 37.0 & 38.54 & 100M \\ 
        ~ & IN1K\(^\ddagger\)  / EN-B7\(^\dagger\) & 15000 & 67.3 & 5.56 & 11.45 & 13.64 & 22.53 & 23.56 & 120M \\ \cmidrule(lr){1-10}
        \multirow{6}{*}{Vidmodex} & MN\(^\ddagger\)  / RN18\(^\dagger\) & 500 & 99.7 & 92.14 & 93.12 & 93.56 & 94.6 & 94.6 & 4M \\ 
        ~ & C10\(^\ddagger\)  / RN18\(^\dagger\) & 1500 & 97.5 & 91.56 & 92.4 & 94.7 & 94.9 & 94.9 & 10M \\ 
        ~ & C100\(^\ddagger\)  / RN34\(^\dagger\) & 3500 & 76.5 & 60.52 & 65.25 & 65.32 & 65.60 & 69.52 & 25M \\ 
        ~ & CT101\(^\ddagger\)  / EN-B7\(^\dagger\) & 8000 & 73.2 & 60.03 & 62.73 & 62.61 & 67.46 & 68.14 & 70M \\ 
        ~ & CT256\(^\ddagger\)  / EN-B7\(^\dagger\) & 10500 & 77.1 & 54.73 & 55.42 & 55.22 & 56.63 & 63.25 & 100M \\ 
        ~ & IN1K\(^\ddagger\)  / EN-B7\(^\dagger\) & 15000 & 67.3 & 32.7 & 37.52 & 40.45 & 43.63 & 48.54 & 120M \\ \bottomrule
    \end{tabular}
    \begin{tablenotes}
        \scriptsize 
        \item{\(^\dagger\)Model Architecture\quad RN-18: ResNet18;\quad RN-34: ResNet34;\quad  EN-B7: EfficientNet-B7}
        \item{\(^\ddagger\)Dataset\quad MN: MNIST;\quad  C10: CIFAR10;\quad  C100: CIFAR100; \quad CT101: Caltech101;\quad  CT256: Caltech256;\quad  IN1K: ImageNet1K}
    \end{tablenotes}    
\end{threeparttable}
}


\end{minipage}
\begin{minipage}{0.495\textwidth}
\resizebox{\linewidth}{!}{
\begin{threeparttable}
    \centering
    \small
    \caption{Comparision of SoftLabel Blackbox Extraction on Video Models across TopK}\label{tab:softlabeltopkVideo}
    \begin{tabular}{cC{2.3cm}C{0.7cm}C{0.7cm}C{0.7cm}C{0.7cm}C{0.7cm}C{0.7cm}C{0.7cm}C{0.95cm}}
    \toprule
        Method & Target Dataset / Victim Model & Victim Train Epochs & Victim Acc. &  \multicolumn{5}{C{5cm}}{Clone Accuracy with given top-k classes in \%}& Query Budget \\ \cmidrule(lr){5-9}
         & & & & 1 & 3 & 5 & 10 & All & \\ \midrule
        \multirow{6}{*}{DFME\cite{truong2021data}} & U11\(^\ddagger\) / VVT\(^\dagger\) & 800 & 84.96 & 49.75 & 50.93 & 52.79 & 55.28 & 55.27 & 70M \\ 
        ~ & U101\(^\ddagger\) / VVT\(^\dagger\) & 2000 & 74.1 & 31.95 & 34.75 & 36.8 & 40.83 & 43.56 & 200M \\ 
        ~ & K400\(^\ddagger\) / SwT\(^\dagger\) & 8000 & 70.8 & 7.52 & 10.42 & 17.24 & 21.56 & 28.49 & 350M \\ 
        ~ & K600\(^\ddagger\) / SwT\(^\dagger\) & 10000 & 68.4 & 2.3 & 3.95 & 7.31 & 12.69 & 18.26 & 420M \\ 
        ~ & SS2\(^\ddagger\) / SwT\(^\dagger\) & 17500 & 61.1 & 0.69 & 1.32 & 4.5 & 8.23 & 11.42 & 500M \\ \cmidrule(lr){1-10}
        \multirow{6}{1.35cm}{\centering DFMS-SL \cite{sanyal2022towards}} & U11\(^\ddagger\) / VVT\(^\dagger\) & 800 & 84.96 & 52.57 & 58.29 & 60.23 & 61.36 & 61.34 & 70M \\ 
        ~ & U101\(^\ddagger\) / VVT\(^\dagger\) & 2000 & 74.1 & 30.6 & 33.81 & 39.37 & 41.4 & 47.53 & 200M \\ 
        ~ & K400\(^\ddagger\) / SwT\(^\dagger\) & 8000 & 70.8 & 20.29 & 20.61 & 21.94 & 27.42 & 34.56 & 350M \\ 
        ~ & K600\(^\ddagger\) / SwT\(^\dagger\) & 10000 & 68.4 & 3.82 & 6.32 & 9.2 & 13.85 & 20.15 & 420M \\ 
        ~ & SS2\(^\ddagger\) / SwT\(^\dagger\) & 17500 & 61.1 & 0.94 & 3.24 & 7.59 & 12.59 & 16.38 & 500M \\ \cmidrule(lr){1-10}
        \multirow{6}{*}{Vidmodex} & U11\(^\ddagger\) / VVT\(^\dagger\) & 800 & 84.96 & 60.14 & 64.53 & 70.24 & 72.71 & 72.64 & 50M \\ 
        ~ & U101\(^\ddagger\) / VVT\(^\dagger\) & 2000 & 74.1 & 48.56 & 53.50 & 57.12 & 61.48 & 68.23 & 200M \\ 
        ~ & K400\(^\ddagger\) / SwT\(^\dagger\) & 8000 & 70.8 & 38.02 & 39.63 & 41.52 & 48.53 & 57.45 & 350M \\ 
        ~ & K600\(^\ddagger\) / SwT\(^\dagger\) & 10000 & 68.4 & 36.91 & 37.59 & 38.31 & 43.51 & 51.62 & 420M \\ 
        ~ & SS2\(^\ddagger\) / SwT\(^\dagger\) & 17500 & 61.1 & 23.94 & 27.51 & 30.41 & 31.94 & 37.63 & 500M \\ \bottomrule 
    \end{tabular}
    \begin{tablenotes}
        \scriptsize 
        \item{\(^\dagger\)Model Architecture\quad VVT: ViViT-B/16x2;\quad SwT: Swin-T;}
        \item{\(^\ddagger\)Dataset\quad U11: UCF-11;\quad U101: UCF-101;\quad K400: Kinetics-400;\quad K600: Kinetics-600;\quad SS2: Something-Something-v2; }
    \end{tablenotes}
\end{threeparttable}
}
\end{minipage}

We note a trend where extraction accuracy increases with the availability of more top-k predictions. This trend is not unique to our method but aligns with findings from DFME \cite{truong2021data} and DFMS-SL \cite{sanyal2022towards}. An interesting observation is that as victim model accuracy decreases, the benefit of having more k predictions becomes more pronounced. This increase can be attributed to the fact that the entropy of the predictions correlates positively with victim accuracy. Thus, by providing more class predictions, we significantly enrich the information available to the system


\subsection{HardLabel}\label{A:hardlabel}


Building on the previous analysis, this section presents a detailed exploration of the hard label scenario, including top-k labels for 1, 3, 5, and 10. Unlike in the soft label scenario, the category 'All' is excluded as it offers no information about the predicted class. This section also details additional metrics such as the architecture of the target victim and clone models, the target dataset, and the number of epochs over which the target is trained. Additionally, the query budget for each experiment and the extraction accuracy are documented. It should be noted that for target datasets comprising only 10 or 11 classes, the top 10 labels effectively represent all classes, rendering this category redundant. The results for the image models are detailed in Table \ref{tab:hardlabeltopkImage}, and those for the video model are shown in Table \ref{tab:hardlabeltopkVideo}.


\begin{minipage}{0.49\linewidth}
\resizebox{\linewidth}{!}{
\begin{threeparttable}
    \centering
    \small
    \caption{Comparision of HardLabel Blackbox Extraction on Image Models across TopK}\label{tab:hardlabeltopkImage}
    \begin{tabular}{cC{2.3cm}C{0.7cm}C{0.7cm}C{0.7cm}C{0.7cm}C{0.7cm}C{0.7cm}C{0.95cm}}
    \toprule
        Method & Target Dataset / Victim Model & Victim Train Epochs & Victim Acc. & \multicolumn{4}{C{4cm}}{Clone Accuracy with given top-k classes in \%} & Query Budget \\ \cmidrule(lr){5-8}
        & & & & 1 & 3 & 5 & 10 & \\ \midrule
        \multirow{6}{*}{DFME\cite{truong2021data}} & MN\(^\ddagger\) / RN18\(^\dagger\) & 500 & 99.7 & 90.1 & 81.25 & 28.64 & - & 4M \\ 
        ~ & C10\(^\ddagger\) / RN18\(^\dagger\) & 1500 & 97.5 & 85.22 & 80.97 & 31.52 & - & 10M \\ 
        ~ & C100\(^\ddagger\) / RN34\(^\dagger\) & 3500 & 76.5 & 52.67 & 54.52 & 52.5 & 13.85 & 25M \\ 
        ~ & CT101\(^\ddagger\) / EN-B7\(^\dagger\) & 8000 & 73.2 & 44.96 & 46.14 & 43.63 & 14.16 & 70M \\ 
        ~ & CT256\(^\ddagger\) / EN-B7\(^\dagger\) & 10500 & 77.1 & 26.71 & 28.5 & 27.73 & 12.72 & 100M \\ 
        ~ & IN1K\(^\ddagger\) / EN-B7\(^\dagger\) & 15000 & 67.3 & 2.8 & 9.53 & 10.42 & 13.21 & 120M \\ \cmidrule(lr){1-9}
        \multirow{6}{1.4cm}{\centering DFMS-HL \cite{sanyal2022towards}} & MN\(^\ddagger\) / RN18\(^\dagger\) & 500 & 99.7 & 86.6 & 86.4 & 23.62 & - & 4M \\ 
        ~ & C10\(^\ddagger\) / RN18\(^\dagger\) & 1500 & 97.5 & 81.29 & 82.52 & 25.74 & - & 10M \\ 
        ~ & C100\(^\ddagger\) / RN34\(^\dagger\) & 3500 & 76.5 & 53.62 & 53.61 & 54.93 & 12.62 & 25M \\ 
        ~ & CT101\(^\ddagger\) / EN-B7\(^\dagger\) & 8000 & 73.2 & 46.7 & 48.52 & 47.61 & 13.84 & 70M \\ 
        ~ & CT256\(^\ddagger\) / EN-B7\(^\dagger\) & 10500 & 77.1 & 32.35 & 32.9 & 31.52 & 13.96 & 100M \\ 
        ~ & IN1K\(^\ddagger\) / EN-B7\(^\dagger\) & 15000 & 67.3 & 5.56 & 18.22 & 19.19 & 15.96 & 120M \\ \cmidrule(lr){1-9}
        \multirow{6}{*}{Vidmodex} & MN\(^\ddagger\) / RN18\(^\dagger\) & 500 & 99.7 & 92.14 & 87.35 & 14.52 & - & 4M \\ 
        ~ & C10\(^\ddagger\) / RN18\(^\dagger\) & 1500 & 97.5 & 91.56 & 86.93 & 17.66 & - & 10M \\ 
        ~ & C100\(^\ddagger\) / RN34\(^\dagger\) & 3500 & 76.5 & 60.52 & 61.52 & 62.44 & 15.63 & 25M \\ 
        ~ & CT101\(^\ddagger\) / EN-B7\(^\dagger\) & 8000 & 73.2 & 60.03 & 58.16 & 57.25 & 9.33 & 70M \\ 
        ~ & CT256\(^\ddagger\) / EN-B7\(^\dagger\) & 10500 & 77.1 & 54.73 & 42.62 & 41.92 & 14.92 & 100M \\ 
        ~ & IN1K\(^\ddagger\) / EN-B7\(^\dagger\) & 15000 & 67.3 & 32.7 & 32.52 & 33.13 & 24.11 & 120M \\ \bottomrule
    \end{tabular}
    \begin{tablenotes}
        \scriptsize 
        \item{\(^\dagger\)Model Architecture\quad RN-18: ResNet18;\quad RN-34: ResNet34;\quad  EN-B7: EfficientNet-B7}
        \item{\(^\ddagger\)Dataset\quad MN: MNIST;\quad  C10: CIFAR10;\quad  C100: CIFAR100; \quad CT101: Caltech101;\quad  CT256: Caltech256;\quad  IN1K: ImageNet1K}
    \end{tablenotes}

\end{threeparttable}
}


\end{minipage}
\begin{minipage}{0.49\linewidth}
\resizebox{\linewidth}{!}{
\begin{threeparttable}
    \centering
    \small
    \caption{Comparision of HardLabel Blackbox Extraction on Video Models across TopK}\label{tab:hardlabeltopkVideo}
    \begin{tabular}{cC{2.3cm}C{0.7cm}C{0.7cm}C{0.7cm}C{0.7cm}C{0.7cm}C{0.7cm}C{0.95cm}}
    \toprule
        Method & Target Dataset / Victim Model & Victim Train Epochs & Victim Acc. & \multicolumn{4}{C{4cm}}{Clone Accuracy with given top-k classes in \%} & Query Budget \\ \cmidrule(lr){5-8}
        & & & & 1 & 3 & 5 & 10 & \\\midrule
        \multirow{6}{*}{DFME\cite{truong2021data}} & U11\(^\ddagger\) / VVT\(^\dagger\) & 800 & 84.96 & 49.75 & 47.15 & 21.94 & - & 70M \\ 
        ~ & U101\(^\ddagger\) / VVT\(^\dagger\) & 2000 & 74.1 & 31.95 & 30.56 & 14.62 & 3.64 & 200M \\ 
        ~ & K400\(^\ddagger\) / SwT\(^\dagger\) & 8000 & 70.8 & 7.52 & 4.62 & 4.72 & 1.92 & 350M \\ 
        ~ & K600\(^\ddagger\) / SwT\(^\dagger\) & 10000 & 68.4 & 2.3 & 2.7 & 2.1 & 0.48 & 420M \\ 
        ~ & SS2\(^\ddagger\) / SwT\(^\dagger\) & 17500 & 61.1 & 0.69 & 0.73 & 1.14 & 1.5 & 500M \\ \cmidrule(lr){1-9}
        \multirow{6}{1.4cm}{\centering DFMS-HL \cite{sanyal2022towards}} & U11\(^\ddagger\) / VVT\(^\dagger\) & 800 & 84.96 & 52.57 & 41.71 & 24.05 & - & 70M \\ 
        ~ & U101\(^\ddagger\) / VVT\(^\dagger\) & 2000 & 74.1 & 30.6 & 31.42 & 25.63 & 11.76 & 200M \\ 
        ~ & K400\(^\ddagger\) / SwT\(^\dagger\) & 8000 & 70.8 & 20.29 & 22.61 & 24.44 & 9.78 & 350M \\ 
        ~ & K600\(^\ddagger\) / SwT\(^\dagger\) & 10000 & 68.4 & 3.82 & 4.15 & 5.15 & 1.36 & 420M \\ 
        ~ & SS2\(^\ddagger\) / SwT\(^\dagger\) & 17500 & 61.1 & 0.94 & 1.14 & 1.74 & 0.93 & 500M \\ \cmidrule(lr){1-9}
        \multirow{6}{*}{Vidmodex} & U11\(^\ddagger\) / VVT\(^\dagger\) & 800 & 84.96 & 60.14 & 52.92 & 36.14 & - & 70M \\ 
        ~ & U101\(^\ddagger\) / VVT\(^\dagger\) & 2000 & 74.1 & 48.56 & 53.62 & 45.39 & 31.62 & 200M \\ 
        ~ & K400\(^\ddagger\) / SwT\(^\dagger\) & 8000 & 70.8 & 38.02 & 35.72 & 32.63 & 21.62 & 350M \\ 
        ~ & K600\(^\ddagger\) / SwT\(^\dagger\) & 10000 & 68.4 & 36.91 & 37.26 & 38.62 & 18.96 & 420M \\ 
        ~ & SS2\(^\ddagger\) / SwT\(^\dagger\) & 17500 & 61.1 & 23.94 & 25.62 & 17.62 & 7.25 & 500M \\ \bottomrule
    \end{tabular}
    \begin{tablenotes}
        \scriptsize 
        \item{\(^\dagger\)Model Architecture\quad VVT: ViViT-B/16x2;\quad SwT: Swin-T;}
        \item{\(^\ddagger\)Dataset\quad U11: UCF-11;\quad U101: UCF-101;\quad K400: Kinetics-400;\quad K600: Kinetics-600;\quad SS2: Something-Something-v2; }
    \end{tablenotes}
    
\end{threeparttable}
}
\end{minipage}

Contrary to the trend observed in the top-k settings for soft labels, there is a decrease in extraction accuracy for hard labels, which can be attributed to a significant increase in the entropy of these model outputs as defined by Eq. \ref{eq:bbharddef}, where probability is equally distributed. This substantial reduction in information from the victim model adversely impacts extraction accuracy, a phenomenon consistent not only with our approach but also observed in studies by DFME \cite{truong2021data} and DFMS-HL \cite{sanyal2022towards}. Additionally, the positive correlation between extraction accuracy and victim model accuracy is evident in the hard label setting, for reasons discussed earlier.

\subsection{Grey Box extraction}\label{sec:appendixgreyboxresults}

\begin{minipage}{0.49\linewidth}
\resizebox{\linewidth}{!}{
\begin{threeparttable}
    \centering
    \small
    \caption{Comparision of GreyBox Extraction on Image Models}\label{tab:greyboxImage}
    \begin{tabular}{cC{2.3cm}C{1.6cm}C{0.7cm}C{0.7cm}C{0.7cm}C{0.7cm}C{0.95cm}C{0.5cm}}
    \toprule
        Method & Target Dataset / Victim Model & Surrogate / Percentage (\%) & Victim Train Epochs & Victim Acc. & \multicolumn{2}{C{1.7cm}}{Clone Accuracy in \%} & Query Budget & Gen Iters \\ \cmidrule(lr){6-7}
        & & & & & SL & HL &  &  \\ \midrule
        \multirow{5}{*}{DFME\cite{truong2021data}} & C10\(^\ddagger\) / RN18\(^\dagger\) & C100\(^\ddagger\) / 10 & 1500 & 97.5 & 93.95 & 89.22 & 5M & 1K \\ 
        ~ & C100\(^\ddagger\) / RN34\(^\dagger\) & IN1K\(^\ddagger\) / 10 & 3500 & 76.5 & 67.51 & 58.92 & 12.5M & 4K \\ 
        ~ & CT101\(^\ddagger\) / EN-B7\(^\dagger\) & CT256\(^\ddagger\) / 10 & 8000 & 73.2 & 59.96 & 49.11 & 30M & 7K \\ 
        ~ & CT256\(^\ddagger\) / EN-B7\(^\dagger\) & IN1K\(^\ddagger\) / 20 & 10500 & 77.1 & 38.63 & 32.15 & 45M & 10K \\ 
        ~ & IN1K\(^\ddagger\) / EN-B7\(^\dagger\) & IN22K\(^\ddagger\) / 5 & 15000 & 67.3 & 29.25 & 22.43 & 58M & 6K \\ \cmidrule(lr){1-9}
        \multirow{5}{1.5cm}{\centering DFMS-(S/H)L \cite{sanyal2022towards}} & C10\(^\ddagger\) / RN18\(^\dagger\) & C100\(^\ddagger\) / 10 & 1500 & 97.5 & 94.86 & 92.62 & 5M & 1K \\ 
        ~ & C100\(^\ddagger\) / RN34\(^\dagger\) & IN1K\(^\ddagger\) / 10 & 3500 & 76.5 & 69.63 & 58.12 & 12.5M & 4K \\ 
        ~ & CT101\(^\ddagger\) / EN-B7\(^\dagger\) & CT256\(^\ddagger\) / 10 & 8000 & 73.2 & 63.62 & 54.85 & 30M & 7K \\ 
        ~ & CT256\(^\ddagger\) / EN-B7\(^\dagger\) & IN1K\(^\ddagger\) / 20 & 10500 & 77.1 & 49.64 & 47.13 & 45M & 10K \\ 
        ~ & IN1K\(^\ddagger\) / EN-B7\(^\dagger\) & IN22K\(^\ddagger\) / 5 & 15000 & 67.3 & 41.52 & 28.62 & 58M & 6K \\ \cmidrule(lr){1-9}
        \multirow{5}{*}{Vidmodex} & C10\(^\ddagger\) / RN18\(^\dagger\) & C100\(^\ddagger\) / 10 & 1500 & 97.5 & 96.61 & 93.24 & 5M & 1K \\ 
        ~ & C100\(^\ddagger\) / RN34\(^\dagger\) & IN1K\(^\ddagger\) / 10 & 3500 & 76.5 & 73.96 & 68.15 & 12.5M & 4K \\ 
        ~ & CT101\(^\ddagger\) / EN-B7\(^\dagger\) & CT256\(^\ddagger\) / 10 & 8000 & 73.2 & 69.16 & 67.72 & 30M & 7K \\ 
        ~ & CT256\(^\ddagger\) / EN-B7\(^\dagger\) & IN1K\(^\ddagger\) / 20 & 10500 & 77.1 & 71.62 & 61.29 & 45M & 10K \\ 
        ~ & IN1K\(^\ddagger\) / EN-B7\(^\dagger\) & IN22K\(^\ddagger\) / 5 & 15000 & 67.3 & 54.12 & 37.16 & 58M & 6K \\ \cmidrule(lr){1-9}
        \multirow{2}{1.5cm}{\centering BlackBox Dissector\cite{wang2022blackdissector}} & C10\(^\ddagger\) / RN34\(^\dagger\) & IN1K\(^\ddagger\) / 100 & - & 91.56 & 80.47 & - & - & - \\ 
        ~ & CT256\(^\ddagger\) / RN34\(^\dagger\) & IN1K\(^\ddagger\) / 100 & - & 78.4 & 63.61 & - & - & - \\ \bottomrule
    \end{tabular}
    \begin{tablenotes}
        \scriptsize 
        \item{\(^\dagger\)Model Architecture\quad RN-18: ResNet18;\quad RN-34: ResNet34;\quad  EN-B7: EfficientNet-B7}
        \item{\(^\ddagger\)Dataset\quad MN: MNIST;\quad  C10: CIFAR10;\quad  C100: CIFAR100; \quad CT101: Caltech101;\quad  CT256: Caltech256;\quad  IN1K: ImageNet1K;\quad IN22K: ImageNet22K}
    \end{tablenotes}
    
\end{threeparttable}
}


\end{minipage}
\begin{minipage}{0.49\linewidth}
\resizebox{\linewidth}{!}{
\begin{threeparttable}
    \centering
    \small
    \caption{Comparision of GreyBox Extraction on Video Models}\label{tab:greyboxVideo}
    \begin{tabular}{cC{2.3cm}C{1.6cm}C{0.7cm}C{0.7cm}C{0.7cm}C{0.7cm}C{0.95cm}C{0.5cm}}
    \toprule
        Method & Target Dataset / Victim Model & Surrogate Dataset / Percentage (\%) & Victim Train Epochs & Victim Acc. & \multicolumn{2}{C{1.5cm}}{Clone Accuracy in \%} & Query Budget & Gen Iters \\ \cmidrule(lr){6-7}
        & & & & & SL & HL &  &  \\ \midrule
        \multirow{6}{*}{DFME\cite{truong2021data}} & U11\(^\ddagger\) / VVT\(^\dagger\) & U101\(^\ddagger\) / 10 & 800 & 84.96 & 58.75 & 51.72 & 45M & 10K \\ 
        ~ & U101\(^\ddagger\) / VVT\(^\dagger\) & K400\(^\ddagger\) / 20 & 2000 & 74.1 & 52.83 & 42.73 & 60M & 15K \\ 
        ~ & K400\(^\ddagger\) / SwT\(^\dagger\) & K600\(^\ddagger\) / 20 & 8000 & 70.8 & 35.73 & 22.67 & 120M & 20K \\ 
        ~ & K600\(^\ddagger\) / SwT\(^\dagger\) & K700\(^\ddagger\) / 20 & 10000 & 68.4 & 32.62 & 20.63 & 200M & 20K \\ 
        ~ & SS2\(^\ddagger\) / SwT\(^\dagger\) & CHRD\(^\ddagger\) / 10 & 17500 & 61.1 & 28.39 & 18.52 & 250M & 10K \\ \cmidrule(lr){1-9}
        \multirow{6}{1.5cm}{\centering DFMS-(S/H)L \cite{sanyal2022towards}} & U11\(^\ddagger\) / VVT\(^\dagger\) & U101\(^\ddagger\) / 10 & 800 & 84.96 & 69.14 & 61.26 & 45M & 10K \\ 
        ~ & U101\(^\ddagger\) / VVT\(^\dagger\) & K400\(^\ddagger\) / 20 & 2000 & 74.1 & 51.20 & 49.15 & 60M & 15K \\ 
        ~ & K400\(^\ddagger\) / SwT\(^\dagger\) & K600\(^\ddagger\) / 20 & 8000 & 70.8 & 48.39 & 27.26 & 120M & 20K \\ 
        ~ & K600\(^\ddagger\) / SwT\(^\dagger\) & K700\(^\ddagger\) / 20 & 10000 & 68.4 & 43.51 & 28.22 & 200M & 20K \\ 
        ~ & SS2\(^\ddagger\) / SwT\(^\dagger\) & CHRD\(^\ddagger\) / 10 & 17500 & 61.1 & 28.01 & 18.05 & 250M & 10K \\ \cmidrule(lr){1-9}
        \multirow{6}{*}{Vidmodex} & U11\(^\ddagger\) / VVT\(^\dagger\) & U101\(^\ddagger\) / 10 & 800 & 84.96 & 73.82 & 65.25 & 45M & 10K \\ 
        ~ & U101\(^\ddagger\) / VVT\(^\dagger\) & K400\(^\ddagger\) / 20 & 2000 & 74.1 & 69.25 & 53.62 & 60M & 15K \\ 
        ~ & K400\(^\ddagger\) / SwT\(^\dagger\) & K600\(^\ddagger\) / 20 & 8000 & 70.8 & 64.02 & 46.93 & 120M & 20K \\ 
        ~ & K600\(^\ddagger\) / SwT\(^\dagger\) & K700\(^\ddagger\) / 20 & 10000 & 68.4 & 53.72 & 39.42 & 200M & 20K \\ 
        ~ & SS2\(^\ddagger\) / SwT\(^\dagger\) & CHRD\(^\ddagger\) / 10 & 17500 & 61.1 & 42.71 & 27.74 & 250M & 10K \\ \bottomrule
    \end{tabular}
    \begin{tablenotes}
        \scriptsize 
        \item{\(^\dagger\)Model Architecture\quad VVT: ViViT-B/16x2;\quad SwT: Swin-T;}
        \item{\(^\ddagger\)Dataset\quad U11: UCF-11;\quad U101: UCF-101;\quad K400: Kinetics-400;\quad K600: Kinetics-600;\quad SS2: Something-Something-v2;\quad K700: Kinetics-700;\quad CHRD: CHARADES }
    \end{tablenotes}
\end{threeparttable}
}
\end{minipage}


Expanding our analysis, we assess the our framework through grey-box model extraction. Although not our primary focus, this experiment is crucial to demonstrate that the SHAP-based objective does not introduce adverse effects during training. This is particularly important as the generator in the black box setting only employs the divergence between the teacher and student models along with the SHAP objective. But in the grey-box scenario, the generator is trained using GAN-based methods with a subset of the target dataset to capture the distribution of samples. Introducing an additional SHAP-based optimization could potentially cause a significant shift in the distribution of these synthetic samples or the generator might encounter a saddle point in this process, impacting the generation of optimal samples for extraction. 


In our grey box extraction analysis, we concentrate on top-all predictions for the softlabel setting and top-1 labels for the hardlabel setting, as these configurations are commonly used in most grey box approaches and have demonstrated superior performance in our prior assessments. We maintain consistency by presenting the same metrics as in previous experiments, with additional details about the surrogate dataset and the proportion of the subset used to train the Generator. Furthermore, we document the number of generator iterations over this dataset to enhance the value of future comparative studies. The results for Image models are detailed in Table \ref{tab:greyboxImage}, and for Video models in Table \ref{tab:greyboxVideo}.


All previously observed trends are maintained in the grey box extraction, exhibiting no significant performance degradation when compared to other algorithms tested, including those documented in \cite{truong2021data} and \cite{sanyal2022towards}. For image models, we also include performance metrics from \cite{wang2022blackdissector}. Our algorithm consistently matches or exceeds the performance of these referenced experiments in both settings.


\section{Qualitative Analysize}\label{A:qualitative}
\subsection{Activation Atlases to visualize efficacy of objectives}\label{A:activationatlas}

The Activation Atlas \cite{carter2019activation} serves to visualize samples \cite{olah2017feature}, optimizing them to align with specific activations using Eq. \ref{eq:cosineactivation}. This alignment corresponds to a UMAP \cite{mcinnes2018umap} based projection of the activation space. For a given sample \(x\), activation maximization targets a specific objective \(\mathcal{J}(\theta_{f}, x, T)\), achieved through the optimization process outlined in Eq. \ref{eq:optimizeactivation}. Here, \(f\) represents the target model, \(\theta_{f}\) its trained parameters, and \(T\) encompasses random transformations (e.g., Flip, Rotate, RandomCrop) applied to \(x\) to enhance generalization and prevent exploitation of local features.


   

\begin{minipage}{0.28\linewidth}
\begin{equation}
x^* = \underset{x}{\mathrm{argmax}} \ \  \mathcal{J}(\theta_{f}, x, T)  
\label{eq:optimizeactivation}
\end{equation}
\end{minipage}
\begin{minipage}{0.719\linewidth}
\begin{align}
    \frac{(h_{x,y} \cdot v)^{n+1}}{(\|h_{x,y}\|\cdot\|v\|)^{n}}; \quad \text{where:} \label{eq:cosineactivation}& & & &  \\ \notag \\
     \tag*{\(v\) is the targeted direction in the activation space,} \\
    \tag*{\(h\) is the activation of the model from the intermediate sample,} \\
    \tag*{\(h_{x,y}\) is the value of the activation vector at the target position.}
\end{align}
\end{minipage}

To demonstrate the quality of information provided by the loss function, we optimize the sample to minimize the loss and visualize the samples in an Activation Atlas, modifying our objective to the negative of the losses in each method. Eq. \ref{eq:actobjDFME} and Eq. \ref{eq:actobjSHAP} detail the objectives employed for the target class \(c\). We utilize a target model trained on ImageNet1K for further analysis. Unlike black box model extraction, where we have only black box access to the model with no data, here we utilize the parameters and compute gradients while having access to the datasets and their respective activations. To test, we replace the generator in our pipeline with the activation maximization algorithm to generate samples that represent the optimal input space that can be learned with respective loss functions. Complete implementation details are provided in the codebase.\footnote{\href{https://github.com/vidmodex/vidmodex/blob/release-code/vidmodex/tests/activation_atlases_lucent_vidmodex.ipynb}{\texttt{https://github.com/vidmodex/.../activation_atlas.ipynb}}} 

\begin{equation}
\mathcal{J}_{DFME}(\theta_{\mathcal{V}}, \theta_{\mathcal{S}}, c, x, T) =  \mathcal{V}(T(x)|c) + \mathcal{S}(T(x)|c) + \sum \mathcal{V}(T(x)|c) \log \frac{\mathcal{V}(T(x)|c)}{\mathcal{S}(T(x)|c)}
\label{eq:actobjDFME}
\end{equation}
\begin{equation}
\mathcal{J}_{SHAP}(\theta_{\mathcal{V}}, \theta_{\mathcal{S}}, \theta_{\mathcal{P}}, c, x, T) = \mathcal{J}_{DFME}(\theta_{\mathcal{V}}, \theta_{\mathcal{S}}, c, x, T) + \mathbb{E}[\mathcal{P}(s|T(x),c)]
\label{eq:actobjSHAP}
\end{equation}

\subsection{Visualization of Training Samples}\label{A:VizTrainingSample}

\begin{figure}[ht!]
    \centering
    \includegraphics[width=0.7\linewidth]{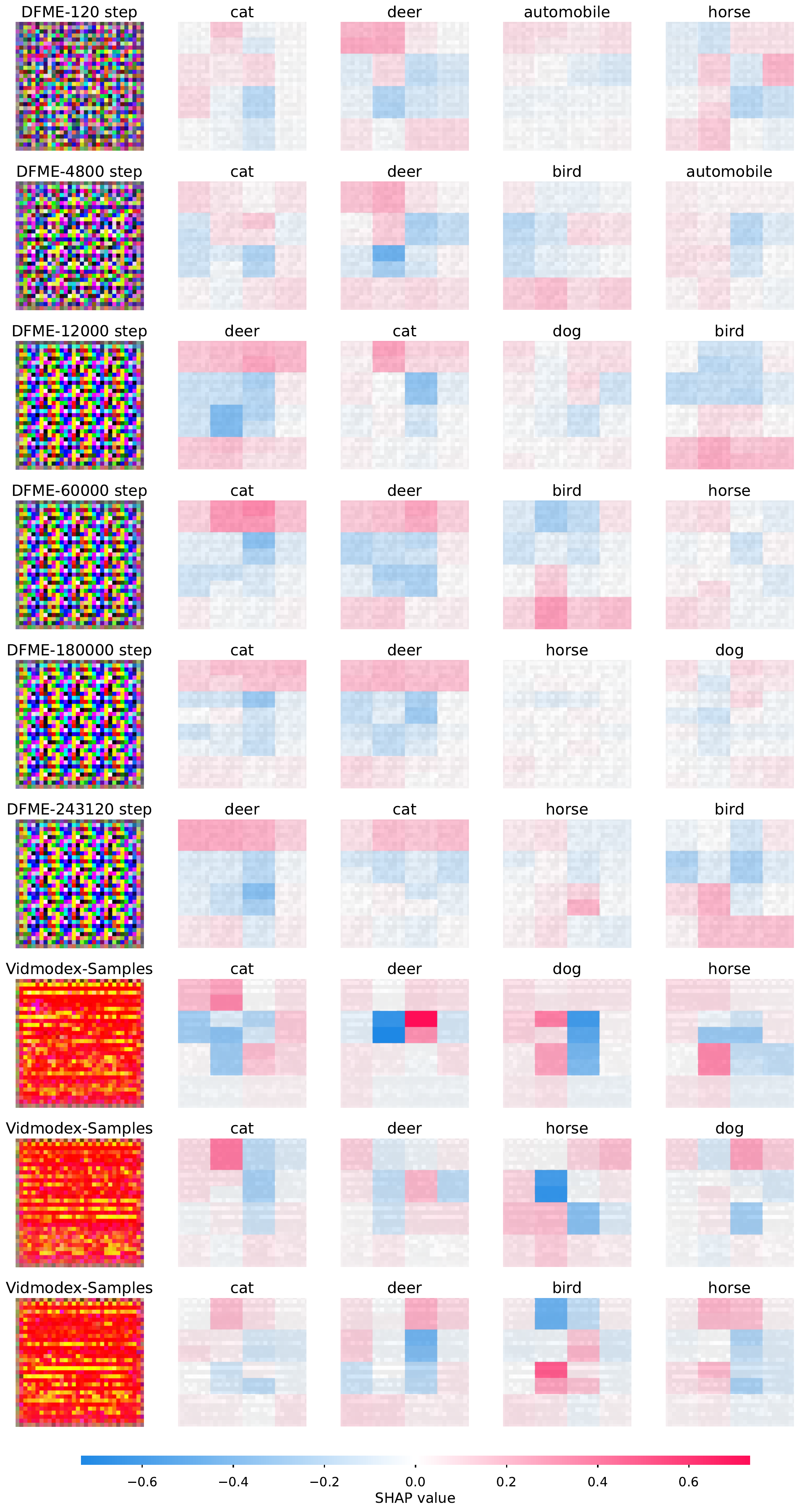}
    \caption{Qualitative analysis of learnt images}\label{fig:qualitativeShap}
\end{figure}

In this section, we showcase samples learned from the DFME approach \cite{truong2021data} and our method across various iteration steps. Our objective is to explore the relationship between SHAP values and image quality within the resulting pipeline. Figure \ref{fig:qualitativeShap} displays samples generated by different algorithms at various timestamps, accompanied by SHAP plots for the top four classes. We have selected CIFAR10 for this analysis to facilitate clearer differentiation among the top four classes.


We present samples for DFME at steps [120, 4800, 12000, 60000, 180000, and 243120]. For our pipeline, we provide three samples from step 2500. Unlike DFME, our generator is conditional, producing all three samples specifically for the cat class. Although SHAP or SHAPley values are not primarily used for learning or improving generator performance, they play a critical role in analyzing sample quality and understanding feature importance within an input. As evident from Figure \ref{fig:qualitativeShap}, samples from the DFME approach display very low SHAP values and lack distinctive features correlating to any particular class. In contrast, samples from Vidmodex, even at intermediate training stages, show many features that are not only optimized for the target class but also exhibit features that negatively correlate with other classes. We restrict our qualitative analysis since all generated samples are non-sensical and cannot be meaningfully interpreted. Therefore, using SHAP values as a sole metric to evaluate these samples is not recommended.



\end{document}